\documentclass{article}
\usepackage{iclr2027_conference,times}
\iclrfinalcopy
\usepackage{amsmath,amssymb}
\usepackage{booktabs,tabularx}
\usepackage[table]{xcolor}
\usepackage{graphicx}
\usepackage{subcaption}
\usepackage{microtype}
\usepackage[scheme=plain,fontset=fandol]{ctex}
\usepackage{tcolorbox}
\tcbuselibrary{breakable}
\usepackage{placeins}
\usepackage{hyperref}
\usepackage{url}

\definecolor{groupblue}{HTML}{EAF0F8}
\definecolor{groupgreen}{HTML}{EAF5EF}
\definecolor{groupcoral}{HTML}{F9EDE8}
\definecolor{agentgreen}{HTML}{DBEEE7}
\definecolor{lightblue}{HTML}{EAF0F8}
\newcommand{\cardiovascauthors}{%
\begin{minipage}[t]{\dimexpr\textwidth-2\tabcolsep\relax}
\centering\normalfont\normalsize
\mbox{Minghao Kong\textsuperscript{\(1,2,\dagger\)}}\quad
\mbox{Muyao Chen\textsuperscript{\(2,\dagger\)}}\quad
\mbox{Jianhao Wu\textsuperscript{\(2\)}}\quad
\mbox{Yuqiang Zhang\textsuperscript{\(2\)}}\quad
\mbox{Xiangbin Meng\textsuperscript{\(2\)}}\quad
\mbox{Zirong Liu\textsuperscript{\(2\)}}\quad
\mbox{Ting Cao\textsuperscript{\(2\)}}\quad
\mbox{Kaida Ning\textsuperscript{\(2\)}}\quad
\mbox{Zhen Song\textsuperscript{\(2,*\)}}\quad
\mbox{Rongjie Wang\textsuperscript{\(2,*\)}}\par
\vspace{0.6em}
\footnotesize
\textsuperscript{1}School of Electronics and Communication Engineering, Sun Yat-sen University, Shenzhen Campus, Shenzhen, China\par
\textsuperscript{2}Department of Network Intelligence, Pengcheng Laboratory, Shenzhen, China\par
\vspace{0.4em}
\textsuperscript{\(\dagger\)}These authors contributed equally to this work.\par
\textsuperscript{*}Corresponding authors: Zhen Song and Rongjie Wang.\par
\href{mailto:songzh01@pcl.ac.cn}{\texttt{songzh01@pcl.ac.cn}}; \href{mailto:wangrj@pcl.ac.cn}{\texttt{wangrj@pcl.ac.cn}}
\end{minipage}%
}
\author{\cardiovascauthors}

\title{CardioVascBench: Prioritized Diagnosis Generation from Real-World Chinese Cardiovascular Records}
\hypersetup{hidelinks,pdftitle={CardioVascBench: Prioritized Diagnosis Generation from Real-World Chinese Cardiovascular Records},
  pdfauthor={Minghao Kong, Muyao Chen, Jianhao Wu, Yuqiang Zhang, Xiangbin Meng, Zirong Liu, Ting Cao, Kaida Ning, Zhen Song, Rongjie Wang}}
\begin{document}
\maketitle
\lhead{Preprint}
\begin{abstract}
Identifying the primary diagnosis alone does not fully characterize a model's diagnosis-generation performance. We introduce CardioVascBench, a retrospective benchmark for prioritized diagnosis generation from real-world Chinese cardiovascular inpatient records. It contains 3,262 de-identified cases, with 2,174 for training and 1,088 for evaluation, linking clinical narratives and textual ECG reports to expert-verified, ordered discharge diagnoses standardized to ICD-10. An English translation of the evaluation set supports aligned cross-lingual comparison. Our evaluation protocol combines terminology normalization and rank-aware scoring to separately assess primary-diagnosis identification, reference-diagnosis coverage, and list precision. Evaluation across 36 systems shows that higher primary-diagnosis accuracy does not necessarily imply higher list precision against discharge references. We also provide CardioKG-Agent, a knowledge-enhanced baseline combining entity normalization, constrained graph retrieval, and guideline evidence. In a staged comparison of three baseline models, supervised fine-tuning plus the agent achieves a 73.5\% higher mean composite score than fine-tuning alone, representing a system-level gain. These results measure retrospective agreement with recorded diagnoses; agreement between metric rankings and physician judgments remains to be validated. CardioVascBench supports research on diagnosis prioritization and knowledge-enhanced generation.
\end{abstract}

\section{Introduction}
\label{sec:introduction}
Generating diagnoses from clinical records requires identifying the primary condition while deciding which additional diagnoses to include and how to order them. In cardiovascular care, this task draws on patient history, examination findings, and ECG reports \citep{wang2024screening,tripathi2022review,yagi2024routine,Braunwald1997}. Large language models (LLMs) have shown promise on medical examinations and clinical questions \citep{singhal2023large,wang2024direct,Nori2023GPT4}, but primary-diagnosis accuracy alone leaves the coverage and precision of the remaining diagnosis list unexamined. CardioVascBench studies these dimensions using real-world Chinese cardiovascular inpatient records.

\begin{figure}[t]
\centering
\includegraphics[width=\linewidth]{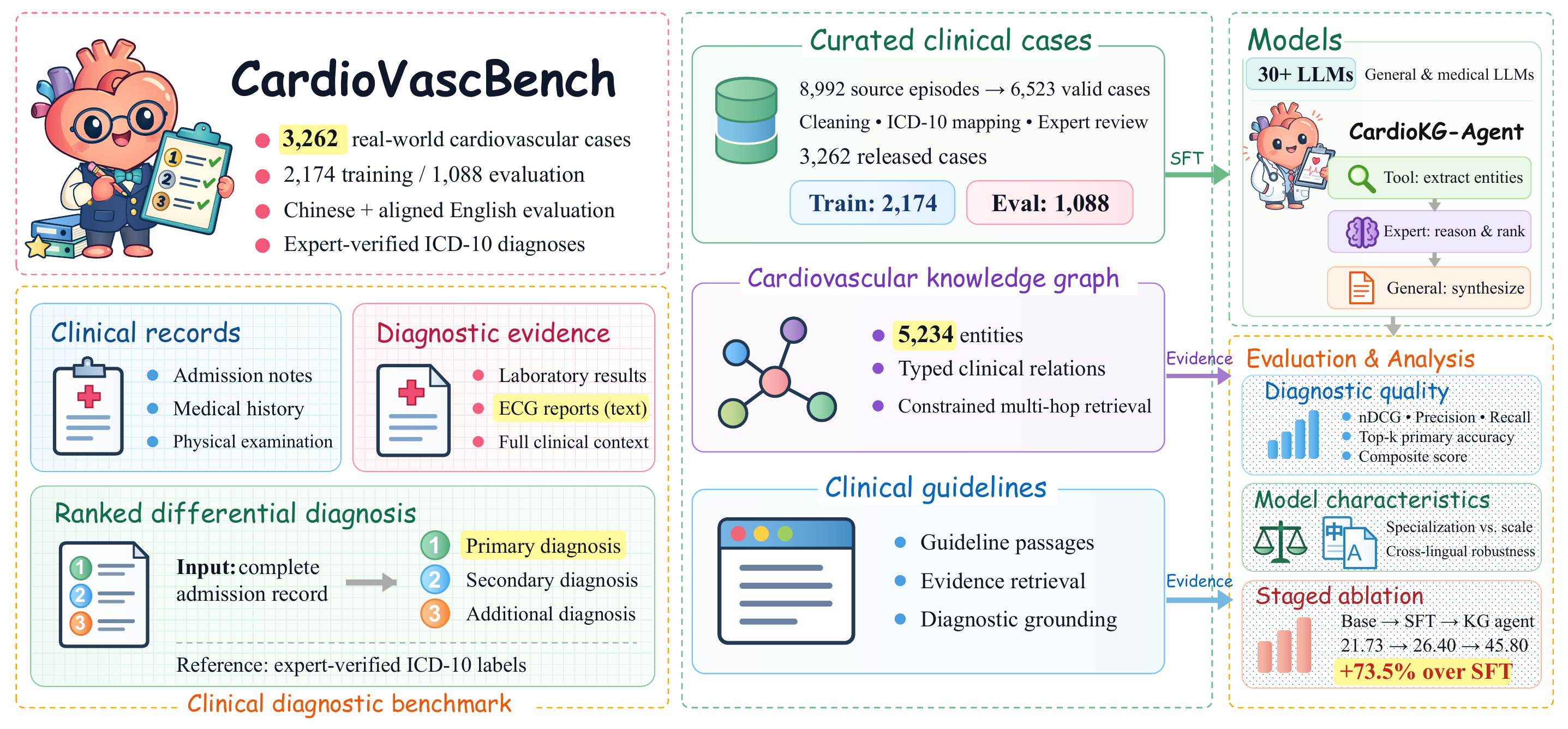}
\caption{\textbf{Overview of CardioVascBench.} Clinical records undergo cleaning, ICD-10 mapping, and expert adjudication to form training and evaluation sets. General and medical-specialized LLMs, together with the CardioKG-Agent baseline, are evaluated on prioritized diagnosis generation. The agent incorporates a cardiovascular knowledge graph and clinical guidelines.}
\label{fig:overview}
\end{figure}

Clinical-record evaluation also requires a clear distinction between agreement with a recorded diagnosis and clinical correctness. Prior studies report limitations in clinical reasoning and free-text recommendations, while tool-augmented systems show benefits on calculation and risk-prediction tasks \citep{hager2024evaluation,williams2024edllm,tordjman2025deepseek,goodell2025clinicalcalc,jin2025agentmd}. These findings motivate evaluating the specific outputs and information access of each system. Here, we assess diagnosis lists against discharge references; this agreement does not by itself establish reasoning fidelity or clinical safety.

We introduce CardioVascBench, a retrospective benchmark of Chinese cardiovascular inpatient records with ordered, ICD-10-standardized discharge diagnoses. Its 3,262 cases include 2,174 training and 1,088 evaluation cases. The benchmark retains variation in clinical narratives, specialist information in ECG reports, and primary--secondary diagnosis structure. A structure-preserving English translation provides an aligned evaluation setting (Appendix~\ref{app:dataset}).

A model may identify the primary diagnosis yet produce an imprecise or incomplete list of additional diagnoses. We therefore evaluate primary-diagnosis identification, coverage, and list precision separately, alongside a position-aware composite score motivated by diagnosis prioritization \citep{Li2025ReEvalMed,Nath2023Primary}. Discharge references support evaluation of prioritized diagnosis generation, but are not exhaustive sets of plausible differential diagnoses. CardioKG-Agent serves as a baseline for studying knowledge augmentation (Figure~\ref{fig:overview}).

We ask three research questions: (RQ1) how well models generate ordered discharge-diagnosis lists from real Chinese cardiovascular records; (RQ2) how primary-diagnosis identification, coverage, and precision differ across models and affect aggregate rankings; and (RQ3) what system-level improvements accompany structured knowledge and guideline evidence.

\noindent\textbf{Our contributions are threefold:} (1) a retrospective benchmark combining real Chinese clinical narratives, cardiovascular information, and ordered diagnosis labels; (2) an evaluation protocol separating primary-diagnosis identification, coverage, and list precision while accounting for terminology and rank; and (3) empirical comparisons of general, medical, and knowledge-enhanced systems, with CardioKG-Agent as a reproducible knowledge-enhanced baseline. The protocol is clinically motivated; its agreement with physician assessment remains an open validation question.

\section{Related Work}
\label{sec:related}
\subsection{Medical reasoning benchmarks and agents}
Established benchmarks assess medical knowledge through exam questions (MedQA \citep{Jin2021MedQA}), biomedical research question answering (PubMedQA \citep{Jin2019PubMedQA}), multitask knowledge tests (MMLU \citep{Hendrycks2021MMLU}), and clinical natural language inference (MedNLI \citep{Romanov2018MedNLI}). More recent evaluations cover broader medical tasks, clinical cases, and automated reasoning assessment \citep{wu2025medsbench,qiu2025medrbench,zhou2026medthink}. AnesSuite evaluates specialty reasoning in anesthesiology at multiple levels of complexity \citep{feng2026anessuite}, while LiveClin uses periodically refreshed case reports to evaluate clinical pathways and reduce contamination risk \citep{wang2026liveclin}. These developments motivate stating precisely which clinical task a benchmark measures.

Clinical-record and signal benchmarks provide complementary perspectives. MIMIC-III supports research with intensive-care records \citep{Johnson2016MIMIC}; Context Clues studies context length and robustness when modeling longitudinal sequences of EHR events \citep{wornow2025contextclues}. Cardiovascular evaluations include ECG-based mortality and risk estimation \citep{Sau2024ECG} and comparisons of ECG foundation models across diverse clinical tasks \citep{almasud2026ecgfms}. CardioVascBench instead evaluates ordered diagnosis generation from Chinese inpatient narratives and ECG reports, with separate measures of primary-diagnosis identification, coverage, and list precision.

Agent evaluations also distinguish answer quality from successful interaction with an environment. MedicalAgentsBench compares reasoning models and agent frameworks on challenging medical questions \citep{Tang2025MedAgents}; MedAgentBench tests clinical tasks in an interactive EHR environment \citep{jiang2025medagentbench}, and MedAgentGym provides executable biomedical data-science tasks with verifiable feedback \citep{xu2026medagentgym}. AMEGA evaluates guideline adherence through open-ended clinical scenarios \citep{fast2024amega}. These settings assess capabilities beyond diagnosis-list agreement: our knowledge-enhanced baseline is evaluated on its final diagnoses, without claiming to measure environment interaction or guideline adherence directly.

\subsection{Multilingual clinical evaluation}
Language coverage remains an important dimension of medical evaluation. Widely used resources such as MIMIC-IV \citep{johnson2023mimic}, PubMedQA, and MedQA include English clinical or biomedical content, while documentation practices and diagnostic terminology vary across languages. MMedC/MMedBench extends multilingual medical QA and rationale generation \citep{qiu2024multilingual}; MedAraBench contributes Arabic medical multiple-choice questions across specialties \citep{abudaoud2026medarabench}. Chinese biomedical understanding and clinical information extraction are supported by CBLUE and unified annotation frameworks \citep{zhang2022cblue,zhu2023unified}. CardioVascBench complements these resources with authentic Chinese cardiovascular inpatient records and ordered diagnosis references; its English translation provides an aligned evaluation setting rather than an independent English patient cohort.

\suppressfloats[t]
\section{CardioVascBench Dataset}
\label{sec:dataset}
\begin{figure}[t]
\centering
\begin{minipage}[c]{0.28\linewidth}
\centering\small
\renewcommand{\arraystretch}{1.35}
\setlength{\tabcolsep}{2pt}
\begin{tabular*}{\linewidth}{@{\extracolsep{\fill}}lrr@{}}
\toprule
\textbf{Dataset} & \textbf{Max} & \textbf{Mean} \\
\midrule
Train-ZH & 10 & 1.12 \\
Eval-ZH & 5 & 1.09 \\
Eval-EN & 5 & 1.09 \\
\bottomrule
\end{tabular*}
\end{minipage}\hfill
\begin{minipage}[c]{0.37\linewidth}
\centering\includegraphics[width=\linewidth]{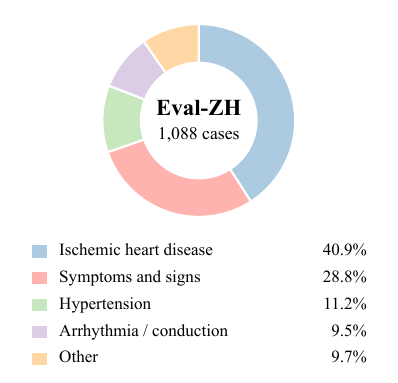}
\end{minipage}\hfill
\begin{minipage}[c]{0.33\linewidth}
\centering\includegraphics[width=\linewidth]{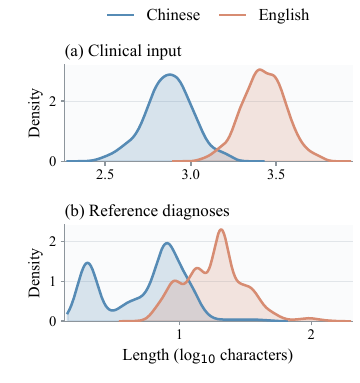}
\end{minipage}
\par\smallskip
\begin{minipage}[t]{0.28\linewidth}
\vspace{0pt}
\captionsetup{font=footnotesize}
\captionof{table}{\textbf{Diagnosis counts.} Maximum and mean entries per case.}
\label{tab:dataset}
\end{minipage}\hfill
\begin{minipage}[t]{0.37\linewidth}
\vspace{0pt}
\captionsetup{font=footnotesize}
\caption{\textbf{Primary diagnoses.} ICD-10 groups in the evaluation set.}
\label{fig:reference-counts}
\end{minipage}\hfill
\begin{minipage}[t]{0.33\linewidth}
\vspace{0pt}
\captionsetup{font=footnotesize}
\caption{\textbf{Text lengths.} Chinese and English evaluation records; non-whitespace characters.}
\label{fig:record-lengths}
\end{minipage}
\end{figure}

\subsection{Dataset overview and clinical task}
CardioVascBench contains 3,262 released, de-identified cardiovascular cases: 2,174 for training and 1,088 for evaluation (Table~\ref{tab:dataset}). Each case represents a clinical episode with admission notes, medical history, physical examinations, laboratory results, ECG reports, and expert-verified ICD-10 diagnoses. The evaluation cases exhibit high completeness for demographics and core history, with an average case length of 554.79 Chinese characters. Detailed statistics and difficulty analyses are provided in Appendix~\ref{app:dataset}. Reference-list cardinality is summarized in Table~\ref{tab:dataset}; Figure~\ref{fig:reference-counts} shows primary-diagnosis groups. Figure~\ref{fig:record-lengths} shows input and reference-diagnosis text lengths for the aligned Chinese and English evaluation sets; these are character counts, not token counts.

The task is \textbf{prioritized diagnosis generation}: given the available narrative and examination reports for an inpatient episode, generate an ordered diagnosis list with the primary diagnosis first, followed by secondary conditions. The reference is the expert-verified hospital discharge-diagnosis list. A plausible differential candidate absent from that list is a non-match under this protocol, but is not necessarily clinically incorrect.

It is important to distinguish between clinical reasoning and coding. In this benchmark, \textbf{ICD-10 codes serve as the standardized label space} for evaluation purposes, enabling rigorous quantitative alignment between model outputs and expert-verified hospital discharge diagnoses. We acknowledge that ICD-10 is a classification system for epidemiological and billing purposes, which may not fully capture the granular nuance of clinical pathophysiology. However, it provides a necessary, fixed vocabulary for automated benchmarking.

\noindent \textbf{Scope and Boundaries.} This benchmark focuses exclusively on the diagnostic phase of clinical care. It \textbf{does not} evaluate downstream clinical tasks such as treatment planning (medication selection, dosage adjustment), procedural recommendations (e.g., PCI vs. CABG), or prognostic assessment. Models are assessed solely on their ability to integrate clinical narratives and examination reports into an accurate and prioritized diagnostic list.

\noindent\textbf{Clinical reasoning foundation.} The protocol is motivated by clinical information integration and diagnosis prioritization \citep{singhal2023large,bowen2006educational,moor2023foundation}. It assesses resulting diagnosis lists against discharge references, without directly evaluating intermediate reasoning. CardioKG-Agent implements a structured baseline workflow; similarity to a clinical workflow does not establish reasoning fidelity.

\subsection{Dataset curation}
The source cohort comprises 8,992 de-identified inpatient episodes from a tertiary cardiology department in China. Records are assembled from admission notes, ECG examination reports, medication and test orders, and structured discharge diagnoses. Cleaning and quality control yield 6,523 valid cases (Figure~\ref{fig:overview}).

DiagnosisICD10Matcher combines semantic matching with SapBERT \citep{liu2021self}, keyword matching, fuzzy matching, and rule-based matching. A multi-LLM voting system assists candidate refinement, while experienced cardiologists make the final adjudications. This process yields the released training set and 1,088 high-confidence evaluation cases. Appendix~\ref{app:dataset} details the processing pipeline and source tables.

\noindent\textbf{Multilingual evaluation.} All evaluation cases are translated into English using a structure-preserving LLM pipeline, preserving the data schema, diagnosis ordering, and gold-standard ICD-10 labels. Disease names are normalized against an official English ICD-10 reference. The English data are used exclusively for evaluation, enabling controlled cross-lingual comparison.

\section{Benchmark Evaluation}
\label{sec:evaluation}
\subsection{Evaluation protocol}
\noindent\textbf{Benchmarked LLMs.} We evaluated 30+ state-of-the-art models, categorizing them into General-Purpose (e.g., GPT-4o \citep{hurst2024gpt}, DeepSeek-V3.2 \citep{liu2025deepseek}) and Medical-Specialized models (e.g., Lingshu-7B \citep{xu2025lingshu}, HuatuoGPT-o1-72B \citep{chen2025towards}). Models were tasked with generating an ordered diagnosis list for each case. Table~\ref{tab:main-results} reports representative models used in our comparisons; Appendix~\ref{app:full-results} preserves the complete results. Evaluation prompts are provided in Appendix~\ref{app:prompts}.

\noindent\textbf{Diagnosis-list metrics.} Diagnostic texts are normalized, and SapBERT matches predictions to reference diagnoses or ICD-10 aliases using a cosine-similarity threshold of $0.80$. Predictions are discounted by rank, while earlier gold diagnoses receive higher importance. We report nDCG, precision, recall, and top-$k$ primary-diagnosis accuracy. The composite score is
\begin{equation}
\mathrm{FinalScore}=\mathrm{nDCG}\,(0.7+0.3\,\mathrm{PrimaryHit})\,\mathrm{Precision}^{\beta},
\label{eq:final-main}
\end{equation}
where $\mathrm{PrimaryHit}$ indicates a match to the primary diagnosis and $\beta$ controls the precision penalty. Appendix~\ref{app:protocol} gives the full definitions and clinical rationale.

\subsection{Knowledge-enhanced baseline}
We include CardioKG-Agent as a knowledge-enhanced baseline for the benchmark. It combines entity normalization, constrained retrieval from a cardiovascular knowledge graph, and guideline evidence to generate an ordered diagnosis list. Architecture and implementation details are provided in Appendix~\ref{app:agent} (Figure~\ref{fig:CardioKG-Agent Architecture}).

\noindent\textbf{Comparison scope.} This baseline has access to external knowledge and guidelines, whereas the base models rely on parametric memory. Comparisons measure system-level performance under different information access and inference budgets; they do not isolate an architectural contribution.

\subsection{Main evaluation results}
\begin{table}[t]
\centering
\caption{\textbf{Main evaluation results on CardioVascBench.} Representative models spanning the groups and comparisons discussed in the text. Values retain the original reporting scale; bold and underlined entries mark the highest and second-highest values per column in this table. Complete results for all 36 systems are in Appendix~\ref{app:full-results}.}
\label{tab:main-results}
\small\setlength{\tabcolsep}{3.2pt}
\begin{tabular*}{\linewidth}{@{\extracolsep{\fill}}lrrrrrr@{}}
\toprule
\textbf{Model} & \shortstack{nDCG\\@10} & \shortstack{Precision\\@10} & \shortstack{Recall\\@10} & Top-1 & Top-3 & Score \\
\midrule
\rowcolor{groupcoral}
\multicolumn{7}{c}{\textbf{Medical-specialized LLMs}} \\
BioMistral-7B & 49.2 & 14.2 & 13.5 & 7.6 & 7.6 & 11.4 \\
HuatuoGPT-o1-7B & 59.7 & 28.7 & 37.1 & 24.3 & 27.1 & 25.4 \\
Lingshu-7B & 61.9 & 32.0 & 40.9 & 24.7 & 27.7 & 28.2 \\
II-Medical-8B & 58.9 & 32.1 & 38.7 & 22.3 & 23.9 & 28.0 \\
HuatuoGPT-o1-72B & 62.2 & 18.3 & 45.0 & 36.6 & 41.3 & 19.5 \\
OpenBioLLM-70B & 58.8 & 30.8 & 34.8 & 14.6 & 16.5 & 25.6 \\
\midrule
\rowcolor{groupblue}
\multicolumn{7}{c}{\textbf{General-purpose open-source LLMs}} \\
Qwen2.5-7B-Instruct & 58.2 & 30.8 & 36.3 & 25.1 & 27.8 & 26.8 \\
Qwen2.5-VL-7B-Instruct & 60.9 & 28.4 & 39.5 & 29.2 & 32.8 & 25.7 \\
Llama3-8B-Instruct & 57.7 & 15.3 & 29.2 & 16.0 & 19.3 & 15.2 \\
Qwen3-8B & 59.4 & 24.8 & 38.5 & 25.9 & 29.4 & 22.9 \\
Llama3-70B-Instruct & 60.0 & 10.7 & 38.7 & 19.3 & 23.1 & 12.7 \\
Deepseek-v3.2 & 61.9 & 16.0 & 43.4 & 33.9 & 40.7 & 17.2 \\
Kimi-K2 & 60.8 & 22.0 & 41.3 & 32.0 & 34.9 & 21.4 \\
\midrule
\rowcolor{groupcoral}
\multicolumn{7}{c}{\textbf{Proprietary LLMs}} \\
GPT-4o & 63.8 & 12.6 & 44.6 & 30.6 & 36.3 & 14.9 \\
GPT-5.2 & \textbf{66.1} & 11.9 & 48.8 & 39.0 & 45.8 & 14.8 \\
Grok-4 & 63.6 & 13.9 & 45.5 & 34.9 & 39.2 & 16.1 \\
Claude-sonnet-4.5 & 65.0 & 11.4 & \textbf{53.0} & \textbf{47.1} & \textbf{55.6} & 15.1 \\
Qwen3-max & 62.8 & 14.7 & 45.3 & 32.5 & 39.0 & 16.7 \\
\midrule
\rowcolor{groupgreen}
\multicolumn{7}{c}{\textbf{Agent-enhanced systems}} \\
Lingshu-7B + Agent & 65.2 & 50.9 & 49.7 & 43.9 & 43.9 & 44.3 \\
HuatuoGPT-o1-7B + Agent & 65.7 & \underline{53.4} & 52.2 & 46.3 & 46.3 & \underline{46.4} \\
OpenBioLLM-70B + Agent & \underline{66.0} & \textbf{53.8} & \underline{52.4} & \underline{46.5} & \underline{46.5} & \textbf{46.7} \\
\bottomrule
\end{tabular*}
\end{table}

\noindent\textbf{Primary diagnosis versus list quality (RQ2).} Claude-sonnet-4.5 has higher Top-1 accuracy and Recall@10 than Lingshu-7B (47.1 versus 24.7, and 53.0 versus 40.9), but lower Precision@10 (11.4 versus 32.0) and final score (15.1 versus 28.2; Table~\ref{tab:main-results}). The ranking reversal illustrates that primary-diagnosis accuracy and reference-based list quality capture different aspects of model performance. Whether the composite ranking better agrees with physician judgment remains to be evaluated.

\noindent\textbf{Specialization and scale.} Smaller medical models can obtain higher composite scores: HuatuoGPT-o1-7B scores 25.4 versus 19.5 for its 72B counterpart. Lingshu-7B and II-Medical-8B score 28.2 and 28.0 versus 15.2 for Llama3-8B-Instruct. These are metric-specific associations across systems with different training and output behavior, rather than evidence that specialization dominates all diagnostic abilities.

\section{Ablation and Analysis}
\label{sec:analysis}
\begin{figure}[t]
\centering
\includegraphics[width=\linewidth]{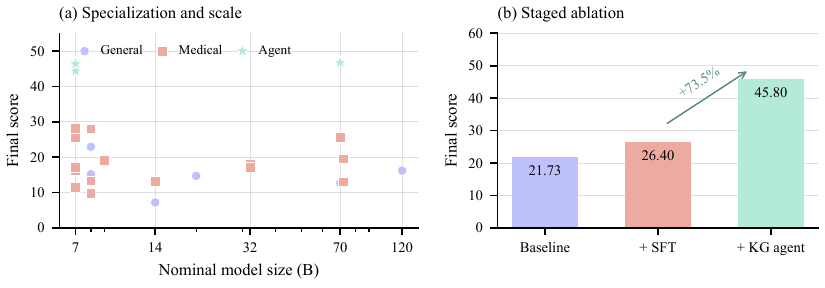}
\caption{\textbf{Model scale and staged ablation.} (a) Final scores for models with a stated parameter count, using the original table values and medical-model classifications. Agent markers use the base model's nominal size. (b) Mean scores for the three ablation configurations. The $+73.5\%$ gain compares SFT with SFT plus the KG agent.}
\label{fig:scale-ablation}
\end{figure}

\subsection{Model characteristics}
\begin{figure}[t]
\centering
\begin{subfigure}[b]{0.48\textwidth}
    \centering
    \includegraphics[width=\textwidth]{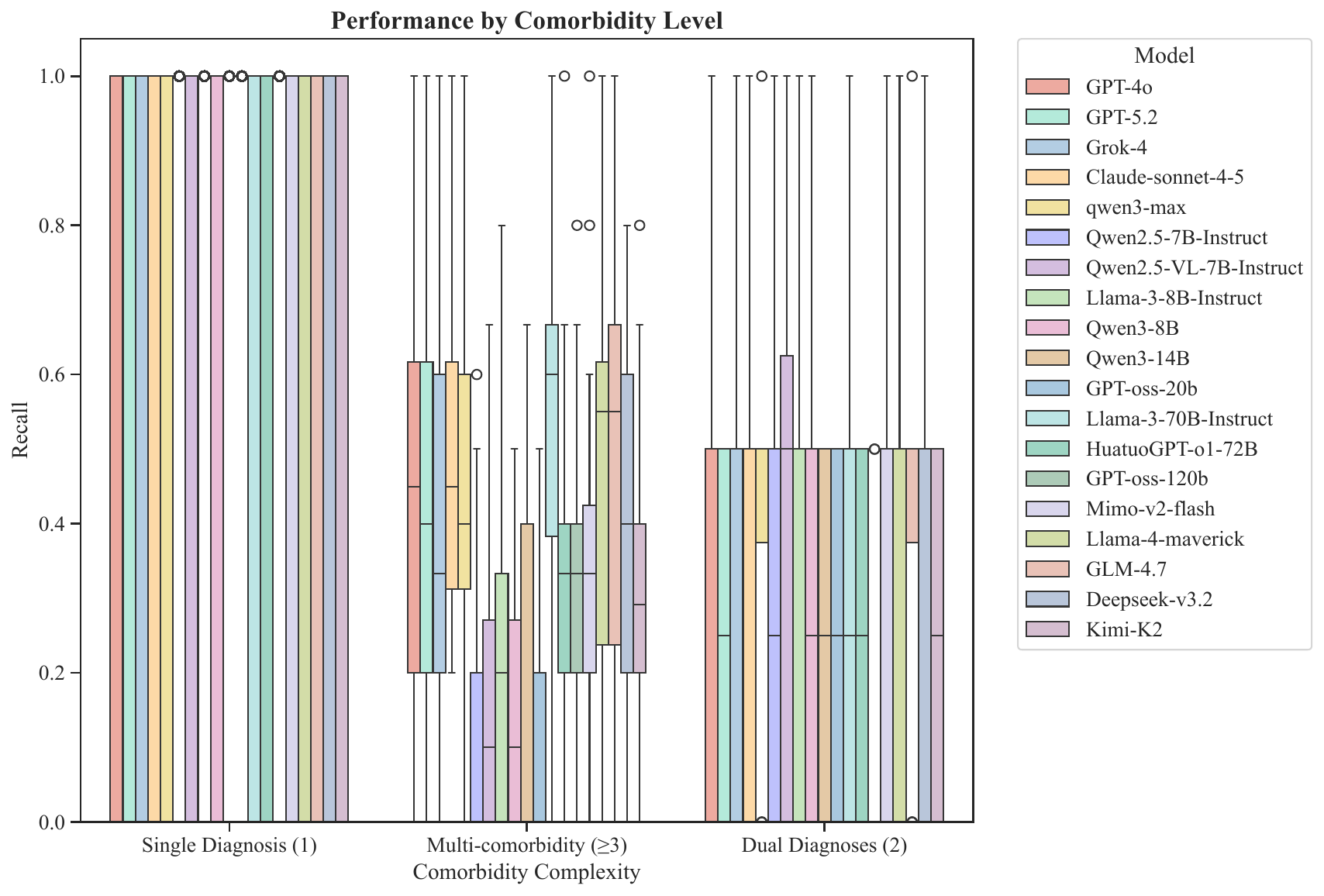}
    \caption{General LLMs}
\end{subfigure}
\hfill
\begin{subfigure}[b]{0.48\textwidth}
    \centering
    \includegraphics[width=\textwidth]{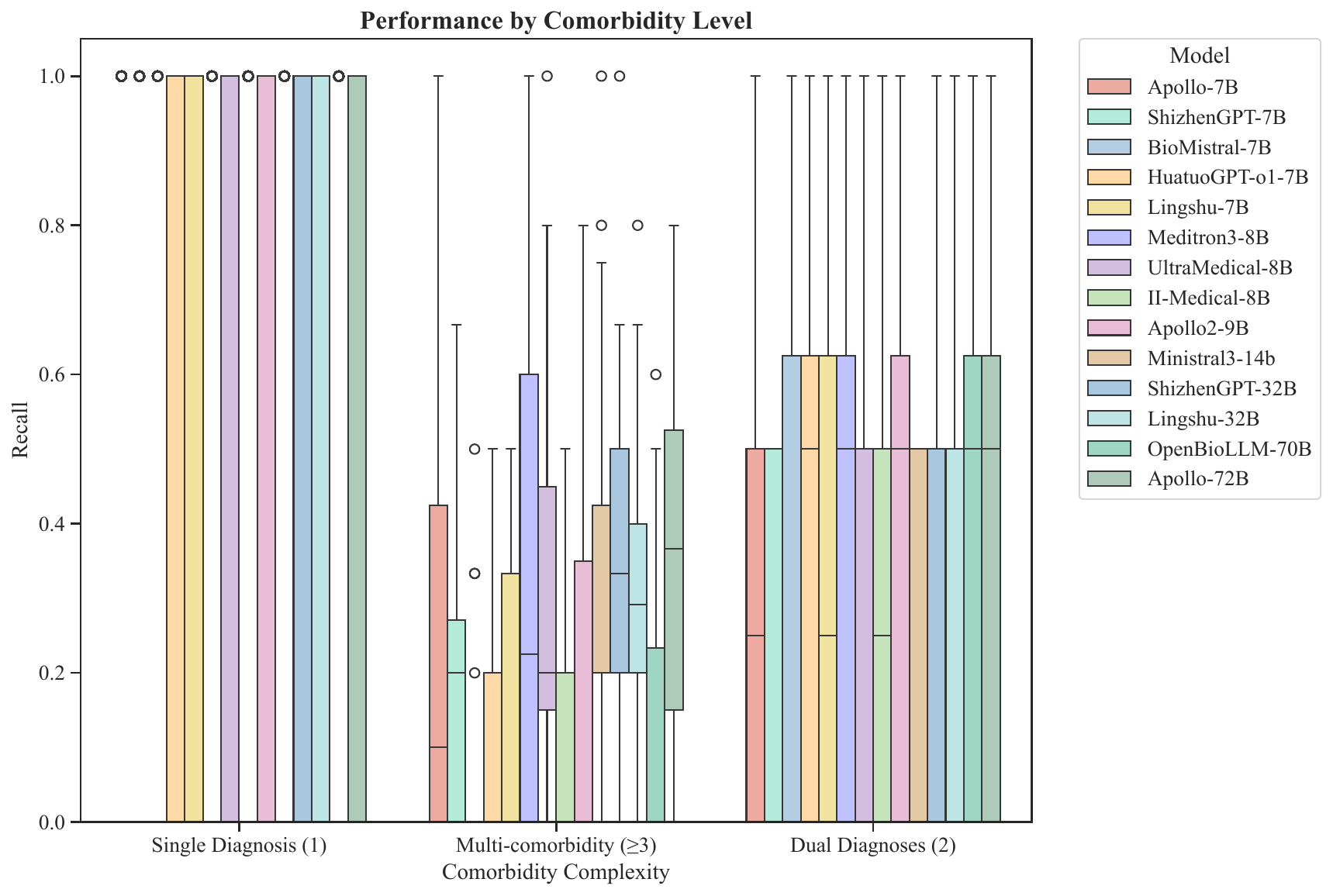}
    \caption{Medical LLMs}
\end{subfigure}
\caption{\textbf{Performance across reported diagnosis-count groups.} General and medical LLMs are compared using the groupings in the original analysis. Stored reference-diagnosis counts do not establish clinical comorbidity prevalence; correspondence between the plotted groups and the prepared evaluation labels requires verification.}
\label{fig:comorbidity_comparison}
\end{figure}
\noindent\textbf{Specialization and scale.} Figure~\ref{fig:scale-ablation}(a) summarizes the relationship between model size and final score. Larger parameter counts do not guarantee better performance on this specialized task. Medical models at smaller scales can outperform much larger general-purpose models, while the agent-enhanced systems obtain the highest composite scores in the reported table.

Figure~\ref{fig:comorbidity_comparison} presents the original comparison across reported diagnosis-count groups. These results are descriptive and do not isolate an effect of domain-specific training. The prepared evaluation set contains only 36 cases with multiple reference entries (Appendix~\ref{app:dataset}), limiting conclusions about multi-diagnosis coverage. The plotted group assignments also require reconciliation with these labels before clinical comorbidity differences can be inferred.

\subsection{Staged ablation}
\begin{table}[t]
\centering
\caption{\textbf{Staged ablation results.} Mean values across the three original baselines; $\Delta$Score is relative to the preceding stage.}
\label{tab:performance}
\small
\setlength{\tabcolsep}{3pt}
\resizebox{\linewidth}{!}{%
\begin{tabular}{lcccccc}
\toprule
Config. & NDCG@10 & Precision@10 & Recall@10 & Top-1 & Score & $\Delta$Score \\
\midrule
Baseline (base LLMs) & 59.70 & 23.30 & 38.17 & 24.53 & 21.73 & -- \\
+SFT & 60.13 & 30.50 & 37.60 & 21.20 & 26.40 & +21.5\% \\
\rowcolor{lightblue}
+SFT + KG\_agent & 65.63 & 52.70 & 51.43 & 45.57 & 45.80 & +73.5\% \\
\bottomrule
\end{tabular}%
}
\end{table}

The staged comparison evaluates Qwen2.5-VL-7B-Instruct, Qwen2.5-7B-Instruct, and Llama3-70B-Instruct on 1,088 cases. The mean final score increases from 21.73 for base LLMs to 26.40 with SFT, and to 45.80 with SFT plus the KG agent (Table~\ref{tab:performance}). The last change is a 73.5\% relative gain over SFT. This is a system-level improvement (RQ3); it does not isolate graph retrieval from guidelines, orchestration, or additional inference budget. Attribution requires the same backbone and comparable budgets across text RAG, graph retrieval, guideline retrieval, and the full system. All reported metrics are in Appendix~\ref{app:analysis}.

\subsection{Misdiagnosis patterns}
We compare Qwen2.5-VL-7B-Instruct, Lingshu-7B, and Lingshu-7B-AGENT using confusion matrices and diagnostic-flow diagrams (Appendix~\ref{app:errors}). The general model displays scattered errors, while the medical model has a tighter diagnostic focus but retains confusion between clinically similar conditions. The agent shows clearer diagonal alignment and more consistent flows toward correct diagnoses. These patterns motivate physician review and confirmatory evidence for high-risk differential diagnoses; they do not establish readiness for autonomous clinical deployment.

\FloatBarrier
\section{Limitations}
\label{sec:limitations}
Discharge references are not exhaustive differential-diagnosis sets, and unmatched candidates are not necessarily hallucinations. Output length, matching rules, and metric weights may influence rankings. Blinded physician assessment, controlled output counts, and weight/matching sensitivity studies are needed to validate the protocol. Historical audits found ECG diagnostic-conclusion fields in all 1,088 evaluation records; absent timestamps and patient linkage prevent certification of a prospective index time or patient-disjoint evaluation. Results are retrospective and episode-level, with input-version provenance requiring verification (Appendix~\ref{app:protocol}). Single-center data limit generalization, and translation can affect cross-lingual comparisons. Agent gains involve different information access and budgets. Neither safety nor reasoning fidelity has been directly evaluated.

\section{Conclusion}
\label{sec:conclusion}
CardioVascBench studies prioritized diagnosis generation from real Chinese cardiovascular records. Its evaluation separates primary-diagnosis identification, coverage, and list precision, exposing differences hidden by a single hit rate. CardioKG-Agent serves as a baseline for knowledge-enhanced generation, with gains over the reported SFT configuration. Physician validation of metric rankings and controlled component comparisons remain necessary next steps.
\label{main-text-end}
\FloatBarrier
\bigskip

\section*{Ethics statement}
This retrospective study was approved by the Institutional Review Board of the First Hospital of Hebei Medical University (Protocol: HDYY-LW-25002). Informed consent was waived because the records were retrospectively collected and de-identified. Patient identifiers were removed before analysis. The evaluated models are for research purposes and are not diagnostic-grade systems. Clinical use requires physician oversight and consideration of patient safety and equitable access.

\section*{Reproducibility statement}
Dataset curation and statistics are described in Section~\ref{sec:dataset} and Appendix~\ref{app:dataset}; the task and evaluation equations are in Section~\ref{sec:evaluation} and Appendix~\ref{app:protocol}. Appendix~\ref{app:agent} describes the agent, Appendix~\ref{app:full-results} reports complete results, and Appendix~\ref{app:prompts} provides the original evaluation prompts. 

\section*{Data and code availability}
The evaluation scripts, agent implementation, and data processing pipelines are available on GitHub at \url{https://github.com/muyaoc/CardioVascBench}. The dataset can be downloaded from OpenI at \url{https://openi.pcl.ac.cn/datasets/detail/cindyc/CardioVascBench}.

\section*{Acknowledgements}
This work is supported by the Major Key Project of Pengcheng Laboratory (PCL2025A12-3) and the National Natural Science Foundation for Young Scholars of China (No. 62103206). This work is also supported by the China Hebei International Joint Research Center for Structural Heart Disease, the Science and Technology Program of Hebei Province (Grant No. 22377719D), the Hebei Province Finance Department Project (Grant No. LS202101), the Hebei Province Higher Education Science and Technology Research Project (Grant No. CXZX2025030), the Hebei Provincial Government-Funded Clinical Talent Project (Grant No. ZF2025062), the Research on Clinical Application of Medical Artificial Intelligence Program of the Hospital Management Institute of the National Health Commission (Grant Nos. YLXX24AIA008 and YLXX24AIA026), the National Natural Science Foundation of China (Grant No. 62202248), and the Science and Technology Planning Project of Guangdong Science and Technology Department under Grant Guangdong Key Laboratory of Advanced IntelliSense Technology (2019B121203006).

\section*{Competing interests}
The authors declare no competing interests.

\section*{AI use statement}
The first draft was written by the authors. GPT-5.2 was used for copy-editing (grammar, spelling) assistance and research queries on related work and references. Prototypes of some icons in the architecture diagram were generated by Gemini-3-Pro, and the authors modified these prototypes into suitable vector graphics. OpenAI Codex assisted with editorial restructuring, LaTeX formatting, figure layout, preparation of an overview schematic, and redrawing plots from the reported tables.

\bibliographystyle{iclr2027_conference}
\bibliography{references}
\clearpage
\appendix
\phantomsection
\label{app:contents}
\begin{center}\large\bfseries Appendix\end{center}
\section*{Contents}
\begingroup
\small
\makeatletter
\@starttoc{apc}
\makeatother
\endgroup
\clearpage
\tcbset{benchmarkprompt/.style={breakable,colback=green!3!white,colframe=green!65!black,colbacktitle=green!65!black,coltitle=white,boxrule=0.5pt,arc=2pt,fonttitle=\small\bfseries,fontupper=\small\ttfamily,left=6pt,right=6pt,top=5pt,bottom=5pt}}

\raggedbottom

\section{Dataset Construction and Statistics}
\addcontentsline{apc}{section}{\protect\numberline{\thesection}Dataset Construction and Statistics}
\label{app:dataset}
\suppressfloats[t]
\subsection{Source cohort and processing}
\addcontentsline{apc}{subsection}{\protect\numberline{\thesubsection}Source cohort and processing}
The dataset comprises 8,992 de-identified cardiovascular inpatient episodes collected from the cardiology department of a tertiary hospital in China. Each episode is indexed by a unique anonymized identifier and assembled from four hospital tables: (1) admission record text, including admission note title, chief complaint, present illness, and past medical history; (2) ECG examination reports (7,362 episodes, 18,463 reports in total), containing examination items, findings, and ECG diagnoses when available; (3) medication and test orders (583,712 orders across all episodes), including order category, type, name, frequency, dose count, and administration route; and (4) structured discharge diagnoses (6,713 episodes, 7,786 diagnosis entries) with ordered ranks. 

\begin{figure}[htbp]
\centering
\includegraphics[width=\textwidth]{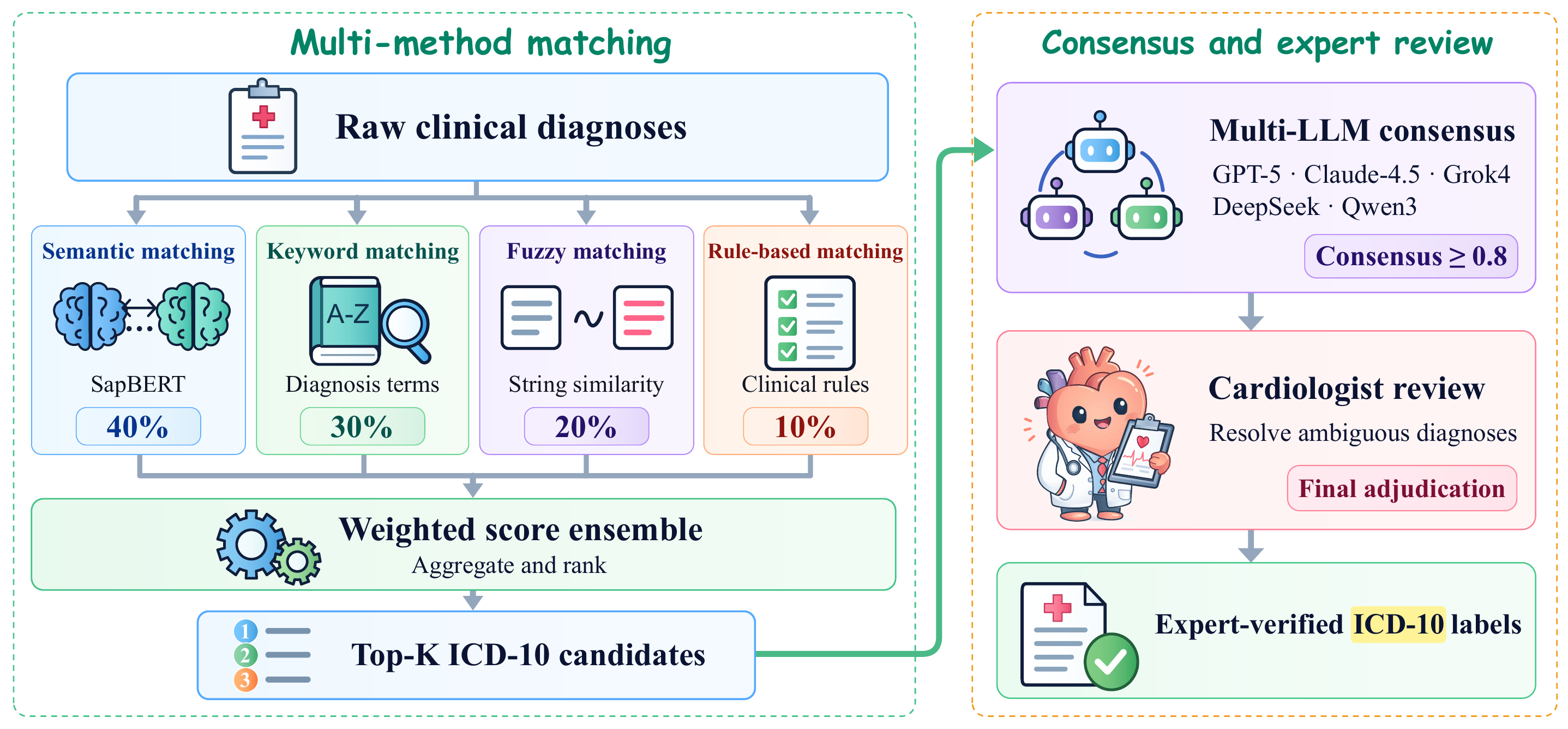}
\caption{\textbf{Multi-Method ICD-10 Matching Architecture}. The framework integrates four distinct matching strategies—Semantic Matching (using SapBERT embeddings), Keyword Matching, Fuzzy Matching, and Rule-Based Matching—to map raw hospital diagnoses to standardized ICD-10 codes. A Multi-LLM voting mechanism (comprising GPT-5, Claude-4.5, Grok4, DeepSeek, and Qwen3) is subsequently employed to resolve ambiguities and refine candidate matches, ensuring high-precision diagnostic labeling for the benchmark. Ultimately, the final adjudications, particularly for controversial or ambiguous diagnoses, were determined by experienced cardiologists.}
\label{fig:Multi-Method ICD-10 Matching Architecture}
\end{figure}

\noindent \textbf{Ethical Approval and Data Governance}. This study was approved by the Institutional Review Board of the First Hospital of Hebei Medical University (Protocol: HDYY-LW-25002). Given the retrospective nature of the study and complete de-identification of patient records, the requirement for informed consent was waived. All data were processed in compliance with relevant data protection regulations. Patient identifiers were removed prior to analysis, and the de-identified dataset underwent additional review to ensure no re-identification risk.

To map raw hospital diagnoses to ICD-10 codes, we developed an ensemble algorithm, DiagnosisICD10Matcher, integrating multi-matching strategies: Semantic Matching (SapBERT \citep{liu2021self} embeddings), Keyword Matching, Fuzzy Matching, and Rule-Based Matching. We further implemented a multi-LLM voting system (GPT-5 \citep{singh2025openai}, Claude-4.5, Grok4 \citep{xai_2025_grok}, DeepSeek, Qwen3 \citep{yang2025qwen3}) to assist in refining candidates ( Figure \ref{fig:Multi-Method ICD-10 Matching Architecture}). From the initial 8,992 episodes, we identified 6,523 valid cases. The results in question shall be determined by experienced doctors and experts. The final dataset comprises 2,174 cases for training and 1,088 high-confidence (exact match) cases for evaluation (total 3,262).

\noindent \textbf{Multilingual (English) Evaluation Data}. To support multilingual evaluation, we further construct an English evaluation dataset derived from the original Chinese benchmark. All evaluation cases are translated into English using a structure-preserving large language model pipeline, ensuring identical data schema, diagnostic ordering, and gold-standard ICD-10 labels.

To reduce translation-induced variability in disease naming, ICD-10 codes in the translated data are aligned with an official English ICD-10 reference, and disease names are normalized accordingly. The resulting English dataset is used exclusively for evaluation, enabling controlled cross-lingual comparison without introducing additional data heterogeneity.

\subsection{Dataset characteristics}
\noindent\textbf{Descriptive-statistics convention.} The training set has 2,174 cases, a maximum of 10 reference diagnosis entries, and a mean of 1.12. Both evaluation-language files contain the same 1,088 case IDs and reference-list cardinalities (maximum 5, mean 1.09). Within the evaluation set, 1,052 cases contain one entry, 12 contain two, and 24 contain three or more. Counts refer to stored reference entries without deduplication, not to inferred comorbidity prevalence. The primary-diagnosis composition in Figure~\ref{fig:reference-counts} uses the first reference entry and mutually exclusive ICD-10 groups: I20--I25 (ischemic heart disease), R00--R99 (symptoms and signs), I10--I15 (hypertension), I44--I49 (arrhythmia/conduction disorders), and all remaining codes (other). These are code-based descriptive groups, not new clinical annotations. For Figure~\ref{fig:record-lengths}, input length sums non-whitespace Unicode characters in string values of the medical-record and ECG-report fields, excluding keys, IDs, and outputs. Reference length includes diagnosis names only, excluding ICD codes and aliases. Curves are Gaussian kernel density estimates of log$_{10}$ lengths. These descriptive statistics use the existing prepared data files; they do not resolve the per-run input provenance discussed in Appendix~\ref{app:protocol}.

\addcontentsline{apc}{subsection}{\protect\numberline{\thesubsection}Dataset characteristics}
CardioVascBench contains 3,262 released, de-identified cardiovascular cases: 2,174 training cases and 1,088 evaluation cases. They were selected from 8,992 source episodes and 6,523 cases retained after cleaning. The records combine clinical narratives and ECG reports with ICD-10-standardized discharge references. Section~\ref{sec:dataset} describes the cleaning, standardization, and adjudication process.

The evaluation set has reported completeness of $100\%$ for demographics and core history, with an average case length of 554.79 Chinese characters. Of its 1,088 cases, 24 ($2.2\%$) contain three or more stored reference entries; these counts describe reference-list cardinality rather than clinical comorbidity prevalence. The diagnosis-frequency summaries and input-characteristic analyses are shown below. Their source populations and input versions should be distinguished from the prepared evaluation-set statistics above.

\begin{figure}[htbp]
\centering
\includegraphics[width=0.70\textwidth, trim=0 0 0 0, clip]{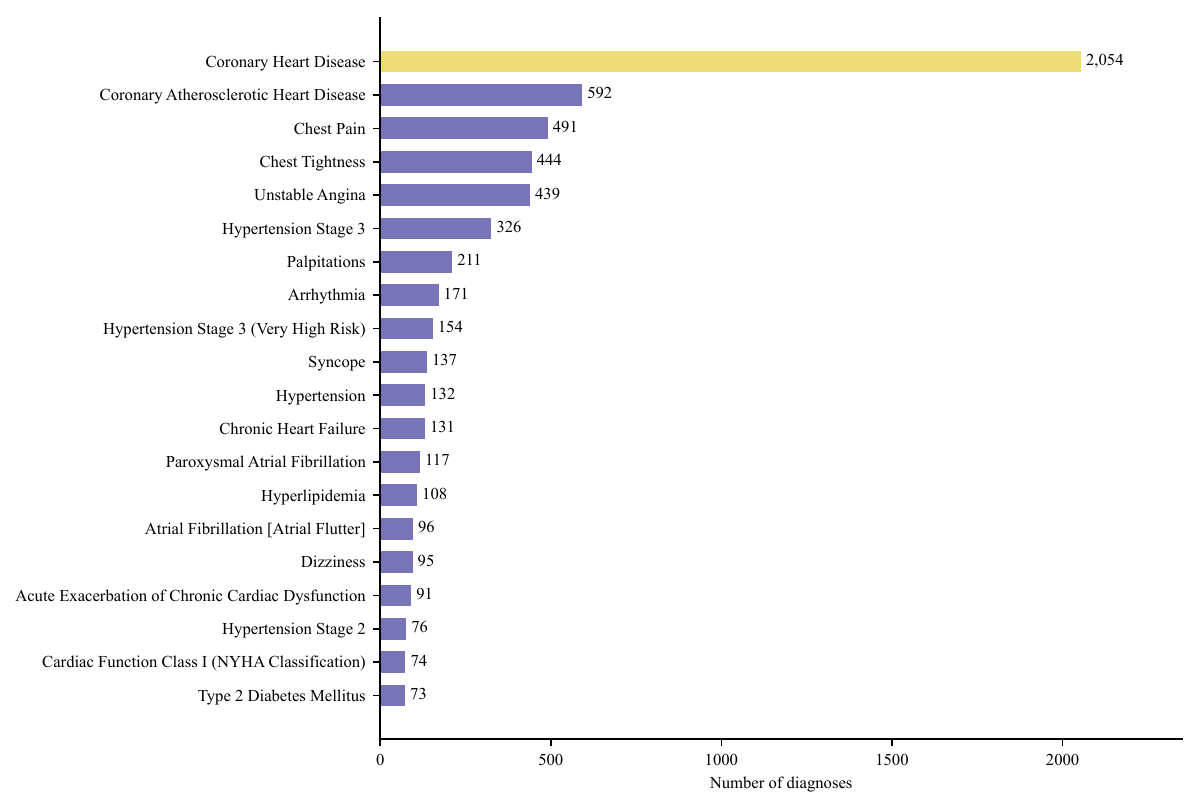}
\caption{\textbf{Diagnosis distribution nature style.} The CardioVascBench is characterized by a high concentration of common cardiovascular diseases and a significant long-tail of infrequent diagnoses. This skewed distribution provides a rigorous testbed for evaluating model reliability, as it requires the agent to navigate both data-rich "head" scenarios and data-sparse "tail" cases, the latter being where traditional parametric LLMs often fail due to knowledge gaps.}
\label{fig:diagnosis distribution nature style}
\end{figure}

We present the detailed distribution of the top 20 diagnoses in our study population (Table \ref{tab:Distribution of Top 20 Diagnoses in the Study Population}) and visualize the characteristic long-tail disease distribution of the CardioVascBench dataset (Figure \ref{fig:diagnosis distribution nature style}). These data illustrate the complexity and real-world nature of the clinical scenarios included in our benchmark.

\begin{table}[htbp]
\centering
\caption{Distribution of Top 20 Diagnoses in the Study Population}
\label{tab:Distribution of Top 20 Diagnoses in the Study Population}
\begin{tabular}{clcc}
\toprule
\textbf{Rank} & \textbf{Diagnosis Name} & \textbf{Count} & \textbf{Proportion} \\
\midrule
1 & Coronary Heart Disease & 2,054 & 26.4\% \\
2 & Coronary Atherosclerotic Heart Disease & 592 & 7.6\% \\
3 & Chest Pain & 491 & 6.3\% \\
4 & Chest Tightness & 444 & 5.7\% \\
5 & Unstable Angina & 439 & 5.6\% \\
6 & Hypertension Stage 3 & 326 & 4.2\% \\
7 & Palpitations & 211 & 2.7\% \\
8 & Arrhythmia & 171 & 2.2\% \\
9 & Hypertension Stage 3 (Very High Risk) & 154 & 2.0\% \\
10 & Syncope & 137 & 1.8\% \\
11 & Hypertension & 132 & 1.7\% \\
12 & Chronic Heart Failure & 131 & 1.7\% \\
13 & Paroxysmal Atrial Fibrillation & 117 & 1.5\% \\
14 & Hyperlipidemia & 108 & 1.4\% \\
15 & Atrial Fibrillation [Atrial Flutter] & 96 & 1.2\% \\
16 & Dizziness & 95 & 1.2\% \\
17 & Acute Exacerbation of Chronic Cardiac Dysfunction & 91 & 1.2\% \\
18 & Hypertension Stage 2 & 76 & 1.0\% \\
19 & Cardiac Function Class I (NYHA Classification) & 74 & 1.0\% \\
20 & Type 2 Diabetes Mellitus & 73 & 0.9\% \\
\bottomrule
\end{tabular}
\end{table}

\subsection{Difficulty Analysis}
\addcontentsline{apc}{subsection}{\protect\numberline{\thesubsection}Difficulty Analysis}

\begin{figure}[htbp]
\centering
\includegraphics[width=0.70\textwidth]{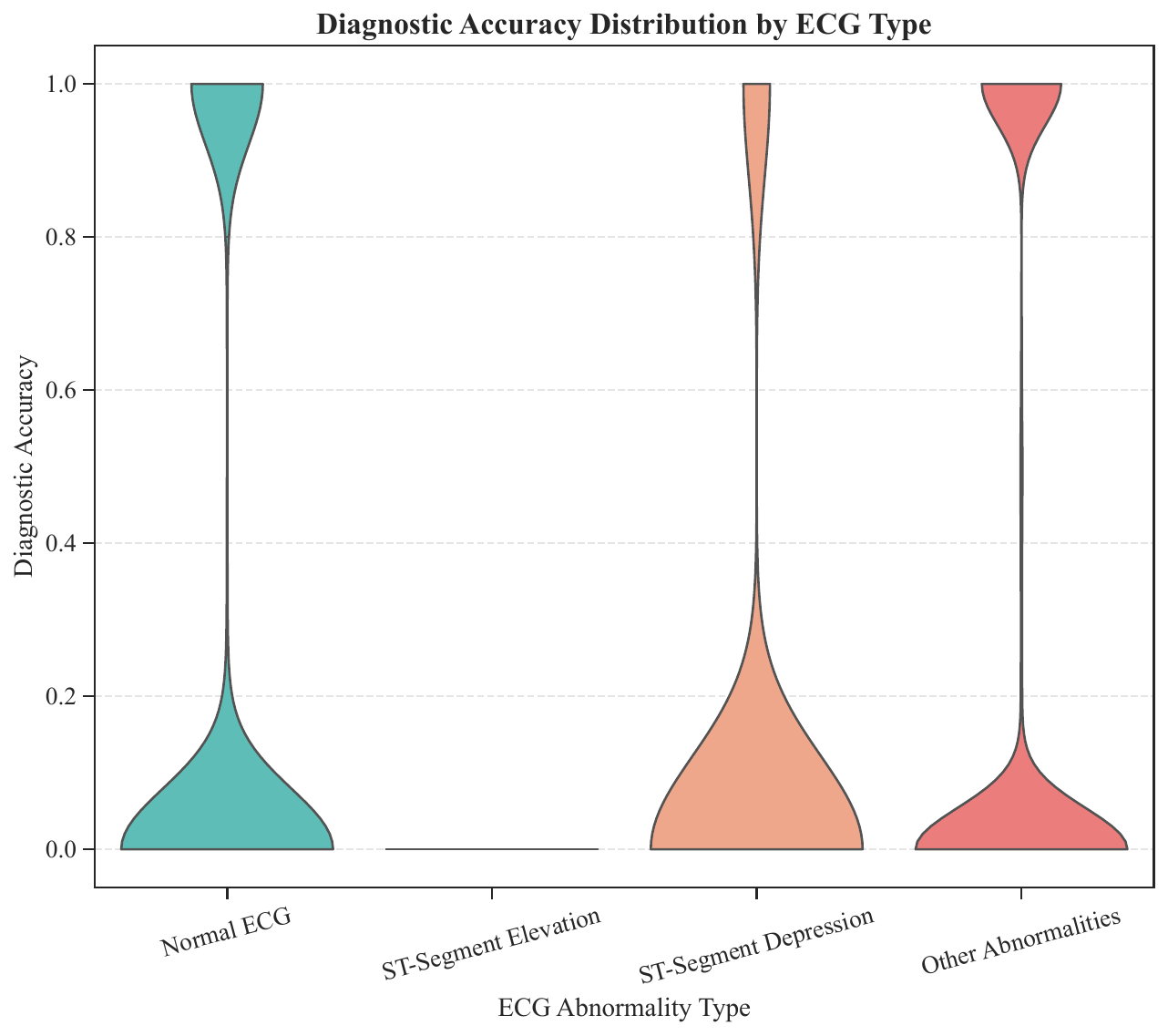}
\caption{\textbf{Reported difficulty by ECG-report availability.} Distributions are shown for the groups in the original analysis. ECG reports are textual inputs; the grouping does not establish a multimodal reasoning requirement.}
\label{fig:ecg_difficulty}
\end{figure}

\begin{figure}[htbp]
\centering
\includegraphics[width=0.70\textwidth]{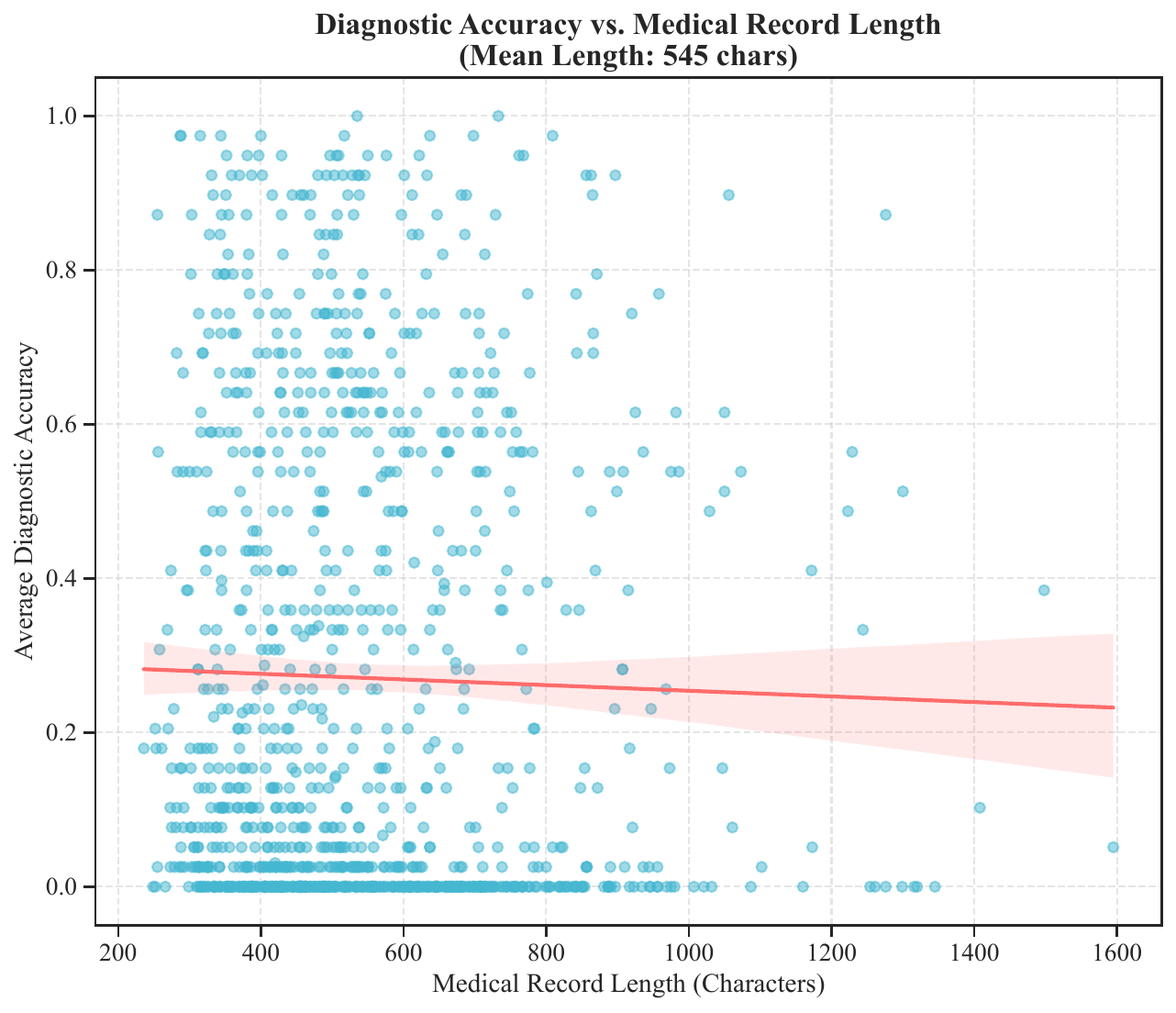}
\caption{\textbf{Clinical-note length and reported difficulty.} Each point relates note length (x-axis) to the reported difficulty score (y-axis). This descriptive plot does not isolate the effect of information volume on diagnostic performance.}
\label{fig:length_difficulty}
\end{figure}

Figures~\ref{fig:ecg_difficulty} and~\ref{fig:length_difficulty} present the reported difficulty scores grouped by ECG-report availability and plotted against clinical-note length. These are descriptive associations: report availability and text length alone do not establish clinical reasoning difficulty.

\FloatBarrier

\FloatBarrier
\section{Clinical Task and Evaluation Protocol}
\addcontentsline{apc}{section}{\protect\numberline{\thesection}Clinical Task and Evaluation Protocol}
\label{app:protocol}

\noindent\textbf{Input-version boundary.} The historical input audit found a nonempty ECG diagnostic-conclusion field in all 1,088 evaluation records. This is not itself proof of temporal leakage, but event timestamps and patient linkage are unavailable to certify temporal eligibility or patient separation. Tables retain historical reported results; their per-run mapping to a frozen input version has not been established in this manuscript. They must not be interpreted as a newly field-filtered or prospective evaluation. Changing inputs requires new predictions.

\noindent\textbf{Unresolved reporting details.} The historical protocol describes up to five generated diagnoses, result tables use @10, and the retained Chinese prompt asks for the final primary diagnosis. These contracts require reconciliation against actual run configurations. Numeric rank-weight and $\beta$ settings also require reporting. Historical prompts and scores are retained without assuming equivalence between configurations.

\subsection{Clinical reasoning foundation}
\addcontentsline{apc}{subsection}{\protect\numberline{\thesubsection}Clinical reasoning foundation}
The protocol is motivated by clinical information integration and diagnosis prioritization \citep{singhal2023large,bowen2006educational,moor2023foundation}. It assesses resulting diagnosis lists against discharge references, without directly evaluating intermediate reasoning. CardioKG-Agent implements a structured baseline workflow; similarity to a clinical workflow does not establish reasoning fidelity.

\subsection{Clinical task definition}
\addcontentsline{apc}{subsection}{\protect\numberline{\thesubsection}Clinical task definition}
The task is \textbf{prioritized diagnosis generation} from the available inpatient narrative and examination reports. Models generate an ordered diagnosis list evaluated against expert-verified discharge diagnoses, with the primary diagnosis first. Plausible differential candidates absent from the discharge reference are non-matches under this protocol, not necessarily clinical errors.

It is important to distinguish between clinical reasoning and coding. In this benchmark, \textbf{ICD-10 codes serve as the standardized label space} for evaluation purposes, enabling rigorous quantitative alignment between model outputs and expert-verified hospital discharge diagnoses. We acknowledge that ICD-10 is a classification system for epidemiological and billing purposes, which may not fully capture the granular nuance of clinical pathophysiology. However, it provides a necessary, fixed vocabulary for automated benchmarking.

\noindent \textbf{Scope and Boundaries.} This benchmark focuses exclusively on the diagnostic phase of clinical care. It \textbf{does not} evaluate downstream clinical tasks such as treatment planning (medication selection, dosage adjustment), procedural recommendations (e.g., PCI vs. CABG), or prognostic assessment. Models are assessed solely on their ability to integrate clinical narratives and examination reports into an accurate and prioritized diagnostic list.

\subsection{Evaluation framework}
\addcontentsline{apc}{subsection}{\protect\numberline{\thesubsection}Evaluation framework}
For each clinical case, the model is provided with the complete medical record, including admission notes, medical history, physical examination findings, laboratory results, and ECG reports. The model is required to generate an ordered list of up to five diagnoses.

To rigorously assess diagnostic performance against expert-annotated gold standards, we adopt a standardized evaluation framework consisting of text normalization, semantic matching, and position-aware ranked metric computation, explicitly designed to reflect real-world clinical diagnostic priorities.

\noindent \textbf{Normalization and Semantic Matching.} 
All diagnostic texts---including model predictions, gold-standard diagnoses, and their corresponding ICD-10 aliases---are first normalized to remove punctuation and unify formatting.

We employ SapBERT, a medical-domain semantic representation model, to compute the semantic similarity between predicted and reference diagnoses. A predicted diagnosis is considered a valid match if the cosine similarity between its embedding and either the gold-standard diagnosis or its ICD-10 alias exceeds a threshold of $0.80$. This semantic matching strategy enables robust evaluation under synonymous or clinically equivalent diagnostic expressions.

\noindent \textbf{Position-Aware Ranked Evaluation.} In clinical practice, diagnoses are not equally important: the primary diagnosis determines treatment decisions and patient management, while secondary diagnoses provide complementary clinical context. To reflect this structure, our evaluation explicitly incorporates rank-dependent weighting on both the gold-standard diagnoses and the model predictions.

\noindent \textbf{Position-Aware Ranked Evaluation.} For a predicted diagnosis at rank j, we apply a logarithmic position discount function:
\begin{equation}
w_{\text{pred}}(j) = \frac{1}{\log_2(j + 1)}
\label{eq:weight}
\end{equation}
This discount assigns substantially higher weight to diagnoses ranked earlier by the model, reflecting the clinical expectation that the most likely diagnosis should be presented first.

\noindent \textbf{Gold-Standard Diagnosis Priority.} Gold standard diagnoses are provided as an ordered list, where the first diagnosis represents the clinically confirmed primary diagnosis, followed by secondary or comorbid conditions. We therefore assign an importance weight to each gold diagnosis based on its position:

\begin{equation}
w_{\text{gt}}(i) = \exp\left(-\lambda (i - 1)\right)
\end{equation}

where $i$ denotes the index of the diagnosis in the gold-standard list, and $\lambda$ controls the strength of prioritization toward the primary diagnosis.

\noindent \textbf{Ranking-based Evaluation.} For a predicted diagnosis $p_j$ and a gold diagnosis $g_i$, the relevance contribution is defined as:
\begin{equation}
\text{rel}(p_j, g_i) = \text{sim}(p_j, g_i) \cdot w_{\text{gt}}(i)
\end{equation}
where $\text{sim}(\cdot)$ is the semantic similarity score computed by SapBERT, and $w_{\text{gt}}(i)$ is a rank-sensitive weight reflecting the clinical importance of the i-th gold diagnosis. This design prioritizes the accurate identification of higher-ranking gold diagnoses, particularly the primary diagnosis.

The Discounted Cumulative Gain (DCG) at rank $K$ is then computed as:
\begin{equation}
\text{DCG@K} = \sum_{j=1}^{K} \max_i \left( \text{rel}(p_j, g_i) \right) \cdot w_{\text{pred}}(j)
\end{equation}
where $w_{\text{pred}}(j)$ is the positional discount factor for the j-th prediction. Thus, semantically correct diagnoses ranked earlier receive greater rewards.

To normalize for cases with different numbers of gold diagnoses and different relevance distributions, we further compute the normalized DCG:
\begin{equation}
\text{nDCG@K} = \frac{\text{DCG@K}}{\text{IDCG@K}}
\end{equation}
where ${\text{IDCG@K}}$ is the maximum achievable DCG under an ideal ordering of the gold diagnoses. This normalization enables a fair comparison across samples with different numbers and rankings of gold diagnoses.

\noindent\textbf{Additional Metrics.} 
To complement nDCG and provide interpretable diagnostic indicators, we also report:
\begin{itemize}
    \item \textbf{Top-$k$ Accuracy} ($k = 1, 2, 3$), measuring whether the primary diagnosis appears within the top-$k$ predictions.
    
    \item \textbf{Precision@$K$} and \textbf{Recall@$K$}, evaluating the correctness and coverage of predicted diagnoses.
    
    \item \textbf{PrimaryHit}, a binary metric specifically designed to assess whether the model successfully identifies the primary diagnosis. PrimaryHit equals $1.0$ if any of the top-$k$ predictions semantically matches the gold-standard primary diagnosis, and $0.0$ otherwise.
\end{itemize}

\noindent\textbf{Final Composite Score.} To summarize the selected ranking and precision dimensions, we define a composite final score:

\begin{equation}
\text{FinalScore} = \text{nDCG} \times (0.7 + 0.3 \times \text{PrimaryHit}) \times \text{Precision}^{\beta}
\end{equation}

where $\beta$ is a hyperparameter controlling the influence of precision.

\noindent\textbf{Rationale and interpretation.} The composite combines ranking, primary-diagnosis recognition, and precision. When $\mathrm{PrimaryHit}=0$, its multiplier is $0.7$; when it is 1, the multiplier is 1. This is a bonus for recognizing the primary diagnosis, not a veto. For $\beta>0$, $\mathrm{Precision}^{\beta}$ penalizes unmatched predictions. Non-matches are defined against discharge references and do not independently establish hallucination or harm. Clinical validity of this scoring choice requires separate assessment.

\noindent\textbf{Rationale for Position-Aware Weighting.} The proposed evaluation framework is motivated by real-world clinical diagnostic practice. In cardiovascular and cerebrovascular diseases, failure to correctly identify or prioritize the primary diagnosis can lead to severe clinical consequences, even if secondary diagnoses are correctly mentioned. Therefore, evaluation metrics that treat all diagnoses equally fail to capture clinically meaningful performance differences.

Rank-aware weighting distinguishes primary and secondary reference diagnoses and discounts later predictions. Component metrics remain necessary to interpret the composite. A clinically motivated weighting scheme does not itself establish agreement with physician judgment.

\subsection{Clinical motivation for prioritization}
\addcontentsline{apc}{subsection}{\protect\numberline{\thesubsection}Clinical motivation for prioritization}
The evaluation measures diagnosis-list agreement after terminology normalization, motivated by primary-diagnosis prioritization \citep{Li2025ReEvalMed,Nath2023Primary}. It does not measure reasoning fidelity or safety. Validation should compare metrics with blinded physician judgments while controlling output count and varying matching thresholds and score weights.

\FloatBarrier
\section{Knowledge-Enhanced Baseline: CardioKG-Agent}
\addcontentsline{apc}{section}{\protect\numberline{\thesection}Knowledge-Enhanced Baseline: CardioKG-Agent}
\label{app:agent}

\begin{figure}[htbp]
\centering
\includegraphics[width=\textwidth]{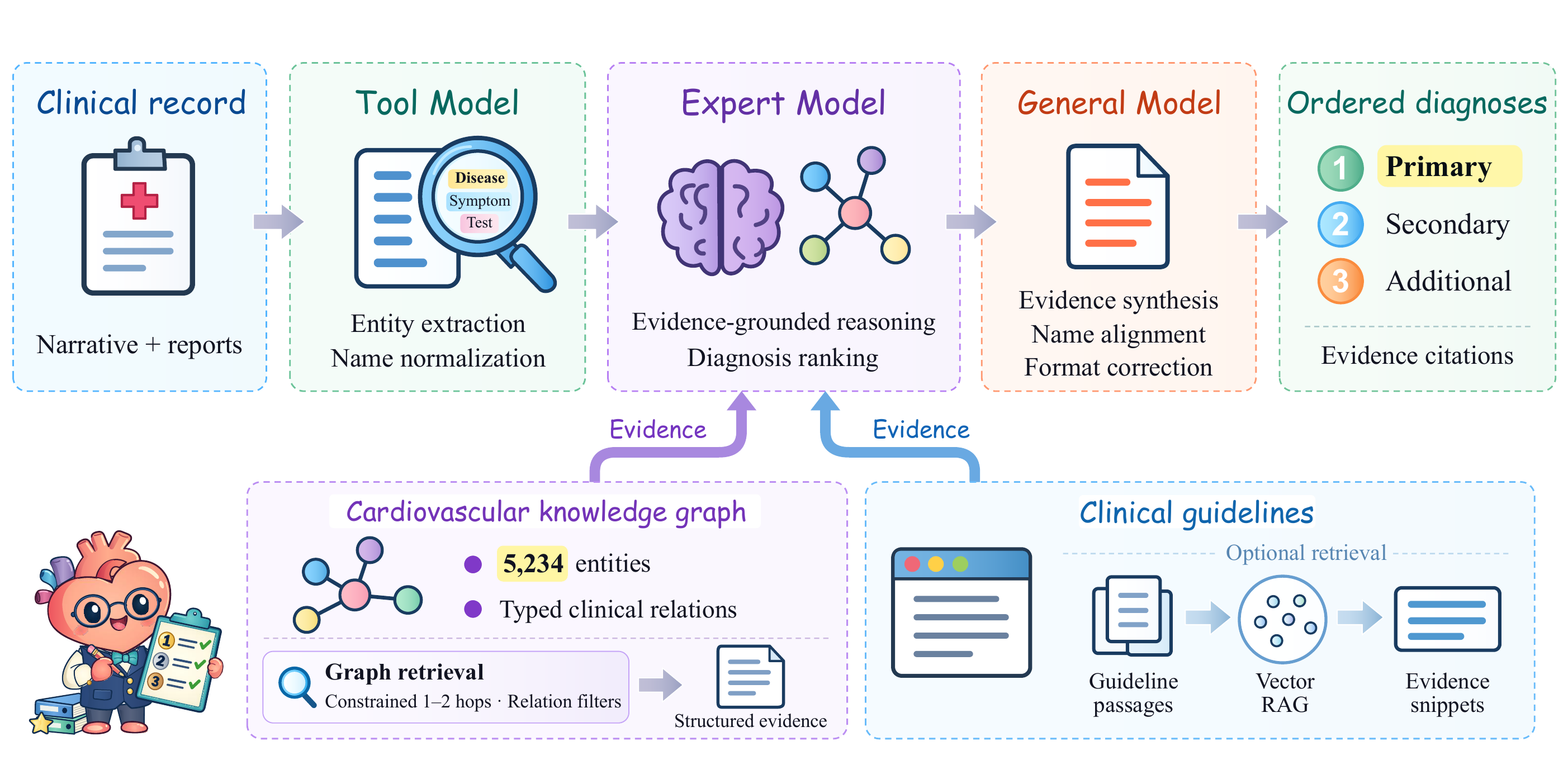}
\caption{\textbf{CardioKG-Agent Architecture.} The diagram illustrates the integration of the Tool Model (extraction), Expert Model (reasoning via KG and Vector RAG), and General Model (synthesis). By combining constrained multi-hop graph traversal with guideline snippet retrieval, the baseline generates an ordered diagnosis list supported by retrieved evidence.}
\label{fig:CardioKG-Agent Architecture}
\end{figure}

We include CardioKG-Agent as a knowledge-enhanced benchmark baseline (Figure \ref{fig:CardioKG-Agent Architecture}). CardioKG-Agent consists of an LLM Core, a cardiovascular knowledge graph (5,234 entities) with a typed schema (e.g., Disease, Symptom, Test, Drug, Treatment, RiskFactor, Complication, Population) and standardized edge labels, and a KG retriever.

Given a full case record, CardioKG-Agent first performs clinical entity extraction and normalization, then retrieves evidence from the KG via constrained multi-hop traversal (e.g., 1–2 hops, relation-type filtered), optionally consults guideline passages for key decision points, and finally synthesizes a prioritized diagnosis list with explicit citations to retrieved evidence. All components are executed in a fixed pipeline to enable reproducibility and ablation.

CardioKG-Agent primarily requires three models: Tool Model, Expert Model, and General Model. The Core Function of the Tool Model is responsible for tasks such as Entity and Extraction. The Core Function of the Expert Model involves activating its own medical knowledge, medical reasoning, and sorting. The General Model is responsible for summarizing the content retrieved from the KG, performing name alignment, and format correction. The models within the Agent perform their respective functions and cooperate with each other, which is an important factor in achieving high scores.
The CardioKG-Agent architecture is explicitly designed to emulate and augment human expert diagnostic reasoning. Like how cardiologists mentally traverse pathophysiological relationships (e.g., 'chest pain \(\rightarrow\) rule out ACS \(\rightarrow\) check troponin and ECG \(\rightarrow\) consider unstable angina vs. NSTEMI'), our agent performs constrained multi-hop traversal through the knowledge graph. The integration of clinical guideline retrieval mirrors how experienced physicians cross-reference authoritative recommendations during complex diagnostic scenarios.

\FloatBarrier
\section{Complete Benchmark Results}
\addcontentsline{apc}{section}{\protect\numberline{\thesection}Complete Benchmark Results}
\label{app:full-results}

This section retains all systems and numerical values from the original manuscript. The main-text table is a representative subset; the table below includes the full proprietary, open-source, medical-specialized, and agent-enhanced results.

\begin{table}[!htbp]
\centering
\caption{\label{tab}\textbf{CardioVascBench Answering Results}}
\footnotesize
\setlength{\tabcolsep}{3pt}

\begin{tabular*}{\linewidth}{@{\extracolsep{\fill}}lrrrrrr@{}}
\hline
\textbf{Model} & \shortstack{\textbf{NDCG}\\\textbf{@10}} & \shortstack{\textbf{Precision}\\\textbf{@10}} & \shortstack{\textbf{Recall}\\\textbf{@10}} & \shortstack{\textbf{Top-1}\\\textbf{Acc}} & \shortstack{\textbf{Top-3}\\\textbf{Acc}} & \shortstack{\textbf{Final}\\\textbf{Score}} \\
\hline
\multicolumn{7}{c}{\textbf{Proprietary Models}} \\
\hline
GPT-4o & 63.8 & 12.6 & 44.6 & 30.6 & 36.3 & 14.9 \\
GPT-5.2 & \textbf{66.1} & 11.9 & 48.8 & 39.0 & 45.8 & 14.8 \\
Grok-4 & 63.6 & 13.9 & 45.5 & 34.9 & 39.2 & 16.1 \\
Claude-sonnet-4.5 & 65.0 & 11.4 & \textbf{53.0} & \textbf{47.1} & \textbf{55.6} & 15.1 \\
Qwen3-max & 62.8 & 14.7 & 45.3 & 32.5 & 39.0 & 16.7 \\
\hline
\multicolumn{7}{c}{\textbf{Open-source Models (<10B)}} \\
\hline
Apollo-7B* & 55.6 & 18.2 & 28.9 & 8.9 & 11.4 & 16.2 \\
ShizhenGPT-7B* & 59.8 & 15.1 & 41.7 & 35.3 & 41.5 & 17.1 \\
BioMistral-7B* & 49.2 & 14.2 & 13.5 & 7.6 & 7.6 & 11.4 \\
HuatuoGPT-o1-7B* & 59.7 & 28.7 & 37.1 & 24.3 & 27.1 & 25.4 \\
Lingshu-7B* & 61.9 & 32.0 & 40.9 & 24.7 & 27.7 & 28.2 \\
Qwen2.5-7B-Instruct & 58.2 & 30.8 & 36.3 & 25.1 & 27.8 & 26.8 \\
Qwen2.5-VL-7B-Instruct & 60.9 & 28.4 & 39.5 & 29.2 & 32.8 & 25.7 \\
Llama3-8B-Instruct & 57.7 & 15.3 & 29.2 & 16.0 & 19.3 & 15.2 \\
Meditron3-8B* & 58.3 & 13.0 & 31.8 & 11.0 & 13.8 & 13.3 \\
Qwen3-8B & 59.4 & 24.8 & 38.5 & 25.9 & 29.4 & 22.9 \\
UltraMedical-8B* & 56.7 & 8.4 & 28.6 & 16.6 & 23.3 & 9.8 \\
II-Medical-8B* & 58.9 & 32.1 & 38.7 & 22.3 & 23.9 & 28.0 \\
Apollo2-9B* & 57.9 & 21.9 & 31.0 & 11.9 & 14.0 & 19.1 \\
\rowcolor{blue!20} \textbf{Lingshu-7B-AGENT} & 65.2 & 50.9 & 49.7 & 43.9 & 43.9 & 44.3 \\
\rowcolor{blue!20} \textbf{HuatuoGPT-o1-7B-AGENT} & \textbf{65.7} & \textbf{53.4} & \textbf{52.2} & \textbf{46.3} & \textbf{46.3} & \textbf{46.4} \\
\hline
\multicolumn{7}{c}{\textbf{Open-source Models (>10B)}} \\
\hline
Ministral3-14B* & 61.1 & 10.7 & 41.4 & 33.7 & 41.4 & 13.2 \\
Qwen3-14B & 36.4 & 6.5 & 19.7 & 12.2 & 15.3 & 7.2 \\
GPT-oss-20B & 58.2 & 16.0 & 26.1 & 11.9 & 15.5 & 14.7 \\
ShizhenGPT-32B* & 62.2 & 16.0 & 44.2 & 39.5 & 47.1 & 17.9 \\
Lingshu-32B* & 61.0 & 16.1 & 39.8 & 31.1 & 36.7 & 17.2 \\
Llama3-70B-Instruct & 60.0 & 10.7 & 38.7 & 19.3 & 23.1 & 12.7 \\
OpenBioLLM-70B* & 58.8 & 30.8 & 34.8 & 14.6 & 16.5 & 25.6 \\
Apollo-72B* & 57.3 & 12.7 & 32.0 & 12.2 & 15.2 & 13.0 \\
HuatuoGPT-o1-72B & 62.2 & 18.3 & 45.0 & 36.6 & 41.3 & 19.5 \\
GPT-oss-120B & 61.6 & 15.1 & 39.3 & 28.2 & 33.6 & 16.2 \\
Mimo-v2-flash & 64.7 & 13.1 & 51.7 & 43.3 & \textbf{50.2} & 16.2 \\
Llama4-maverick & 64.2 & 13.5 & 48.1 & 40.7 & 45.8 & 16.1 \\
GLM-4.7 & 63.3 & 11.1 & 49.9 & 43.7 & 49.1 & 14.4 \\
Deepseek-v3.2 & 61.9 & 16.0 & 43.4 & 33.9 & 40.7 & 17.2 \\
Kimi-K2 & 60.8 & \textbf{22.0} & 41.3 & 32.0 & 34.9 & \textbf{21.4} \\
\rowcolor{blue!20} \textbf{OpenBioLLM-70B-AGENT} & \textbf{66.0} & \textbf{53.8} & \textbf{52.4} & \textbf{46.5} & 46.5 & \textbf{46.7} \\
\hline
\end{tabular*}
\vspace{3pt}
\parbox{0.98\linewidth}{\footnotesize \textbf{Note:} Bold values indicate the best result within each model block for the corresponding metric. `*' denotes medical-specialized models. Blue-shaded rows denote agent-enhanced systems.}
\end{table}

\begin{figure}[htbp]
\centering
\includegraphics[width=\textwidth, trim=0 0 0 0, clip]{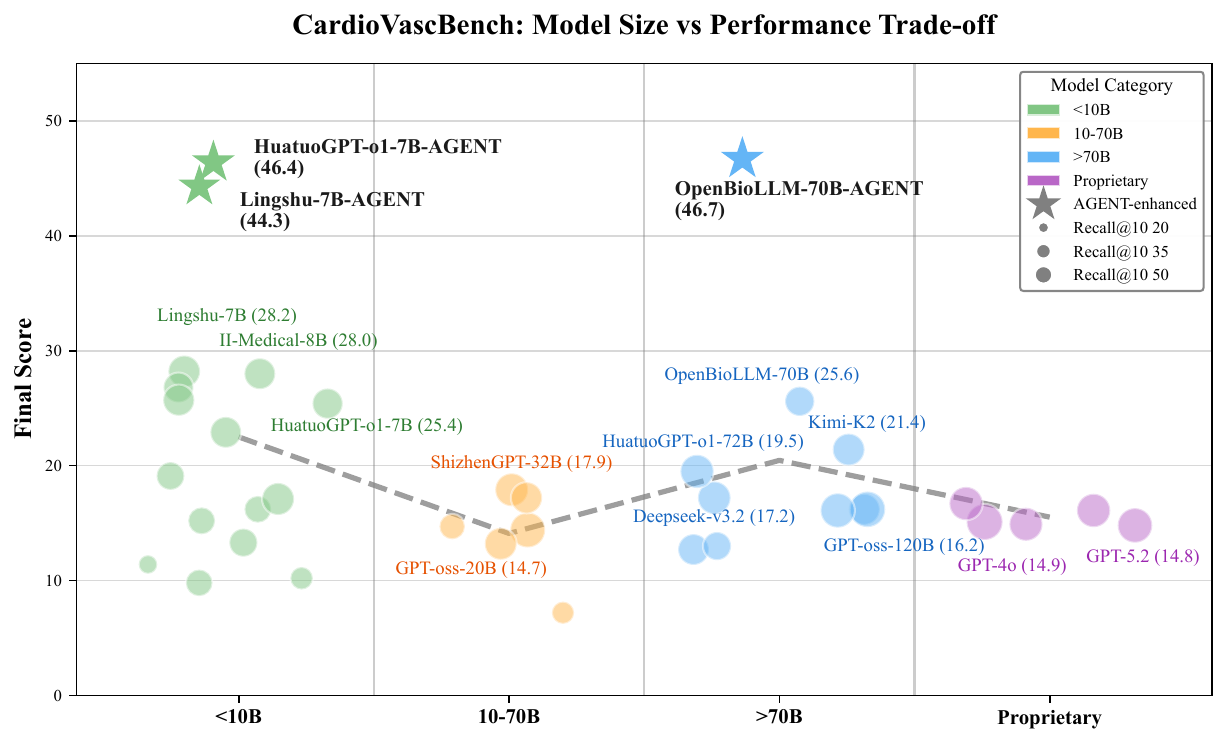}
\caption{\textbf{Composite scores by model size and type.} Agent-enhanced systems obtain higher composite scores than the compared base models under the reported settings. These comparisons reflect system-level differences in knowledge access and inference configuration.}
\label{fig:CardioVascBench scatter final}
\end{figure}

\FloatBarrier
\section{Additional Analyses}
\addcontentsline{apc}{section}{\protect\numberline{\thesection}Additional Analyses}
\label{app:analysis}
\suppressfloats[t]

\subsection{Original ablation visualization}
\addcontentsline{apc}{subsection}{\protect\numberline{\thesubsection}Original ablation visualization}
\begin{figure}[htbp]
\centering
\includegraphics[width=0.6\textwidth, trim=0 0 0 0, clip]{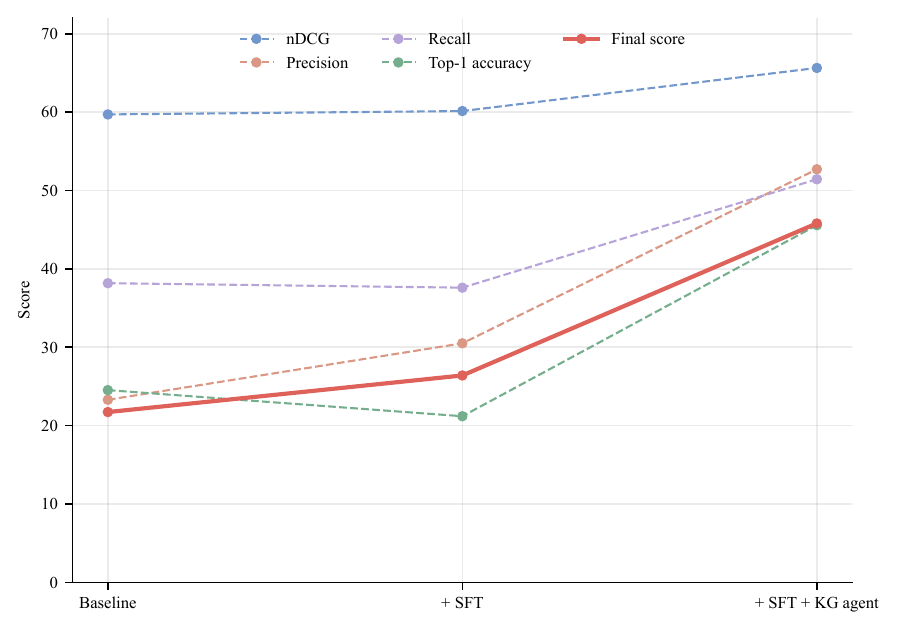}
\caption{\textbf{Ablation Study Results}. Systematic ablation study results on three baselines (Qwen2.5-VL-7B-Instruct, Qwen2.5-7B-Instruct, and Llama3-70B-Instruct) are presented, with values averaged across models (N=1088). The SFT-plus-agent configuration improves the mean score by 73.5\% relative to SFT; this does not isolate graph retrieval.}
\label{fig:Ablation Study Results}
\end{figure}

\subsection{Diagnosis-count group comparison}
\addcontentsline{apc}{subsection}{\protect\numberline{\thesubsection}Diagnosis-count group comparison}
Figure~\ref{fig:comorbidity_comparison} presents the original comparison across reported diagnosis-count groups. These results are descriptive and do not isolate an effect of domain-specific training. The prepared evaluation set contains only 36 cases with multiple reference entries (Appendix~\ref{app:dataset}), limiting conclusions about multi-diagnosis coverage. The plotted group assignments also require reconciliation with these labels before clinical comorbidity differences can be inferred.

\subsection{Misdiagnosis patterns and diagnostic flows}
\begin{figure}[htbp]
\centering
\includegraphics[width=0.32\textwidth]{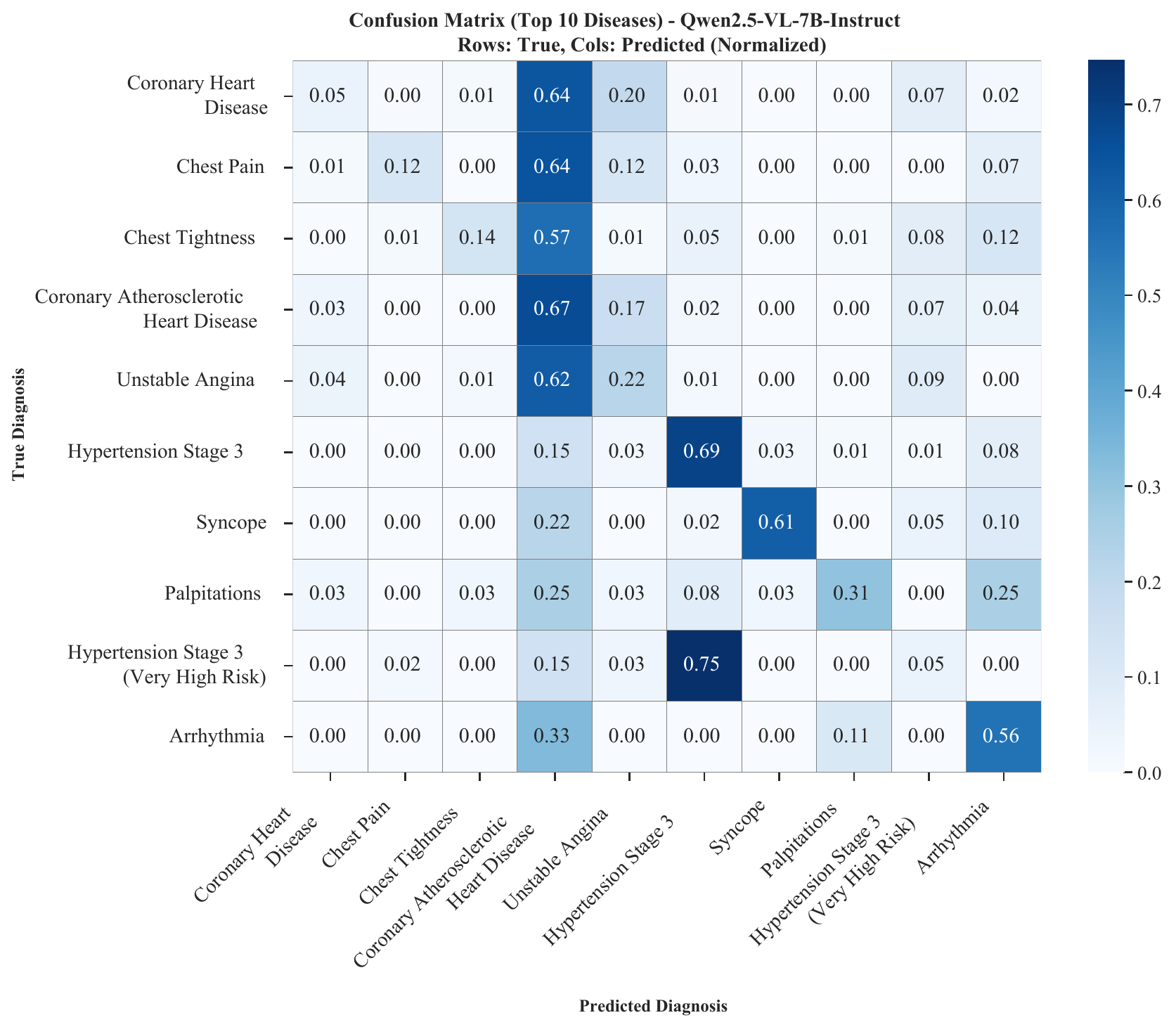}
\hfill
\includegraphics[width=0.32\textwidth]{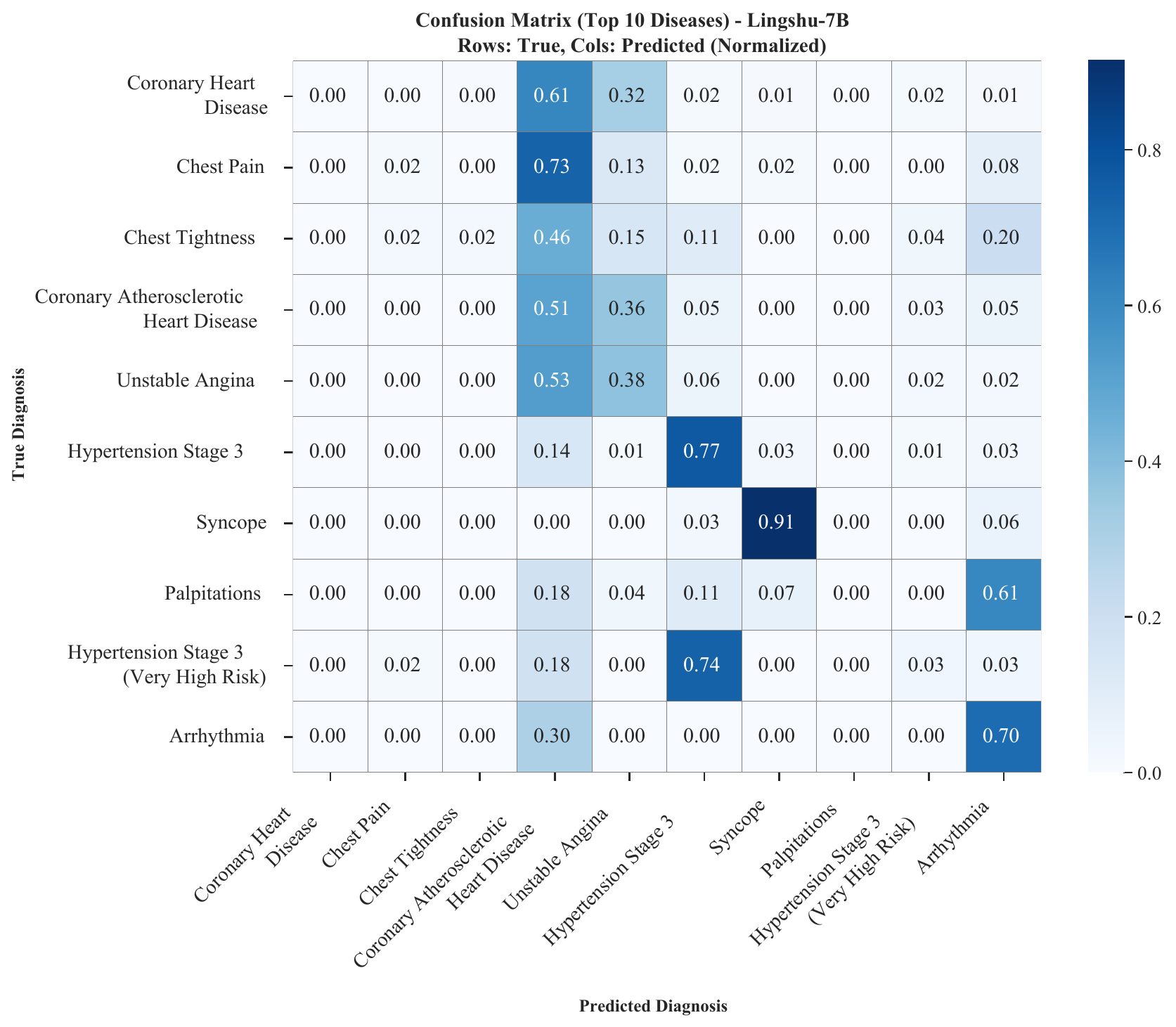}
\hfill
\includegraphics[width=0.32\textwidth]{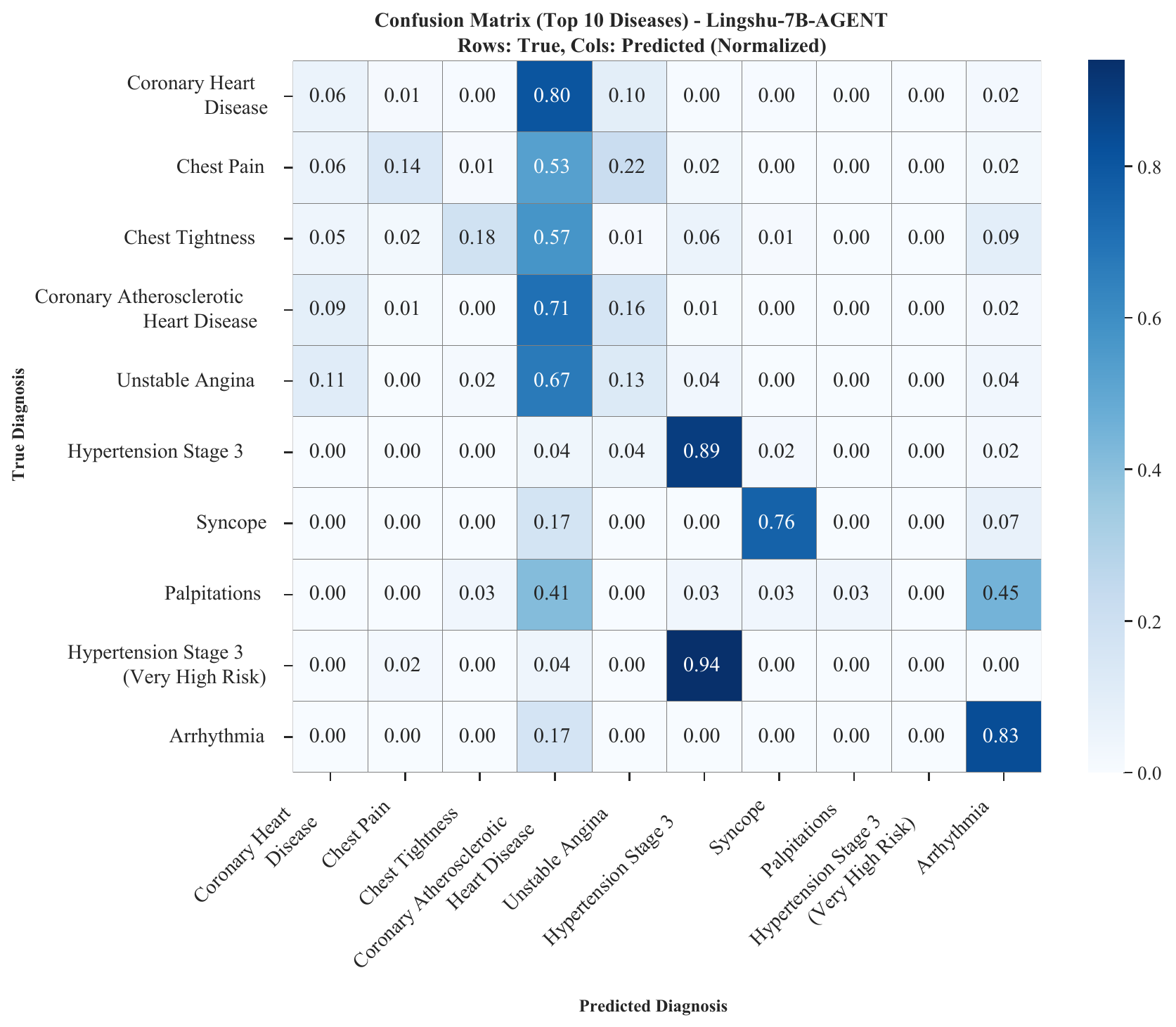}
\caption{\textbf{Horizontal Comparison of Misdiagnosis Patterns (Confusion Matrices)}. From left to right: Qwen2.5-VL-7B-Instruct, Lingshu-7B, and Lingshu-7B-AGENT. The general model exhibits dispersed error patterns, while the agent shows fewer off-diagonal label confusions in these examples.}
\label{fig:confusion_matrices_comparison}
\end{figure}

\begin{figure}[htbp]
\centering
\includegraphics[width=0.32\textwidth]{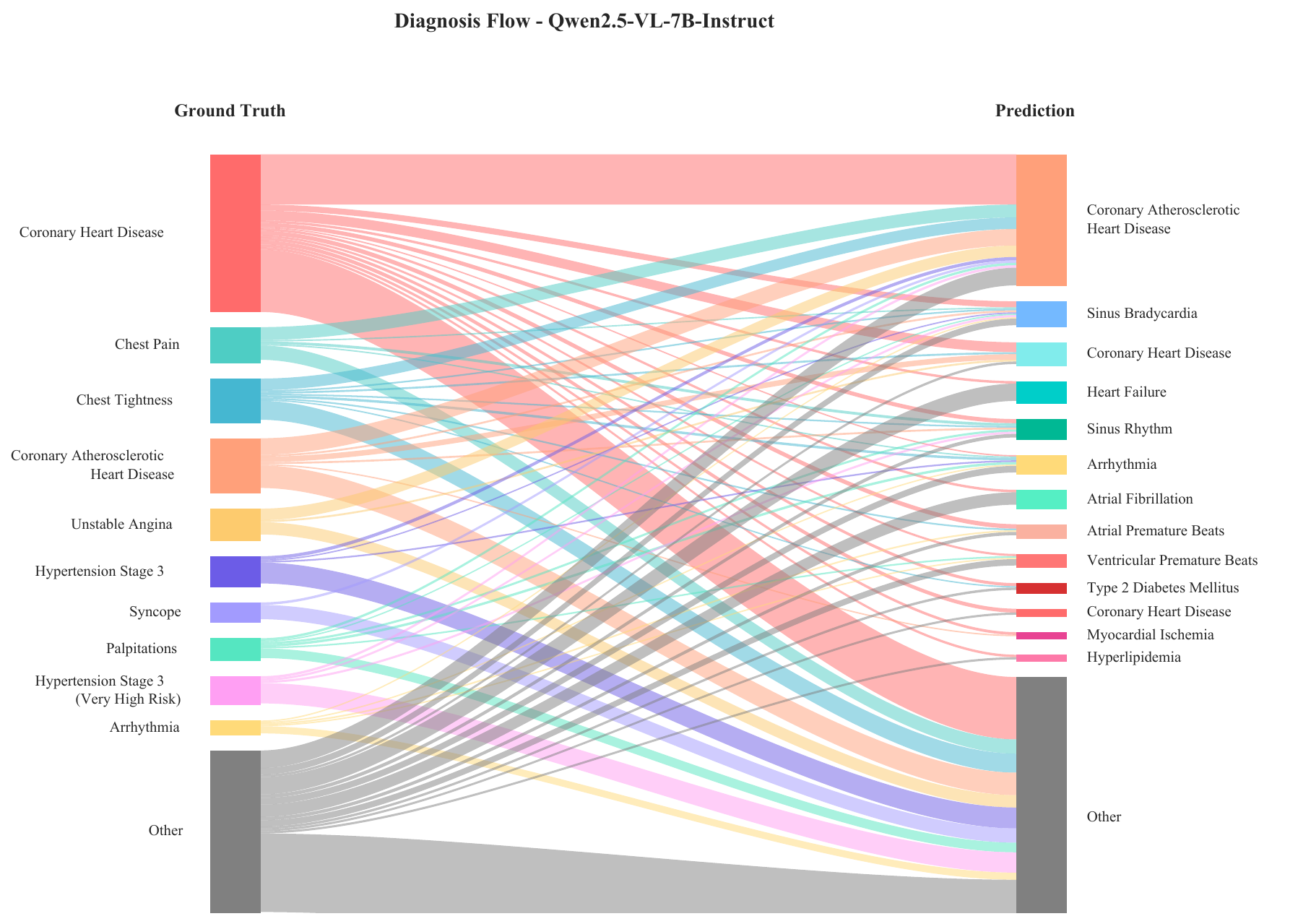}
\hfill
\includegraphics[width=0.32\textwidth]{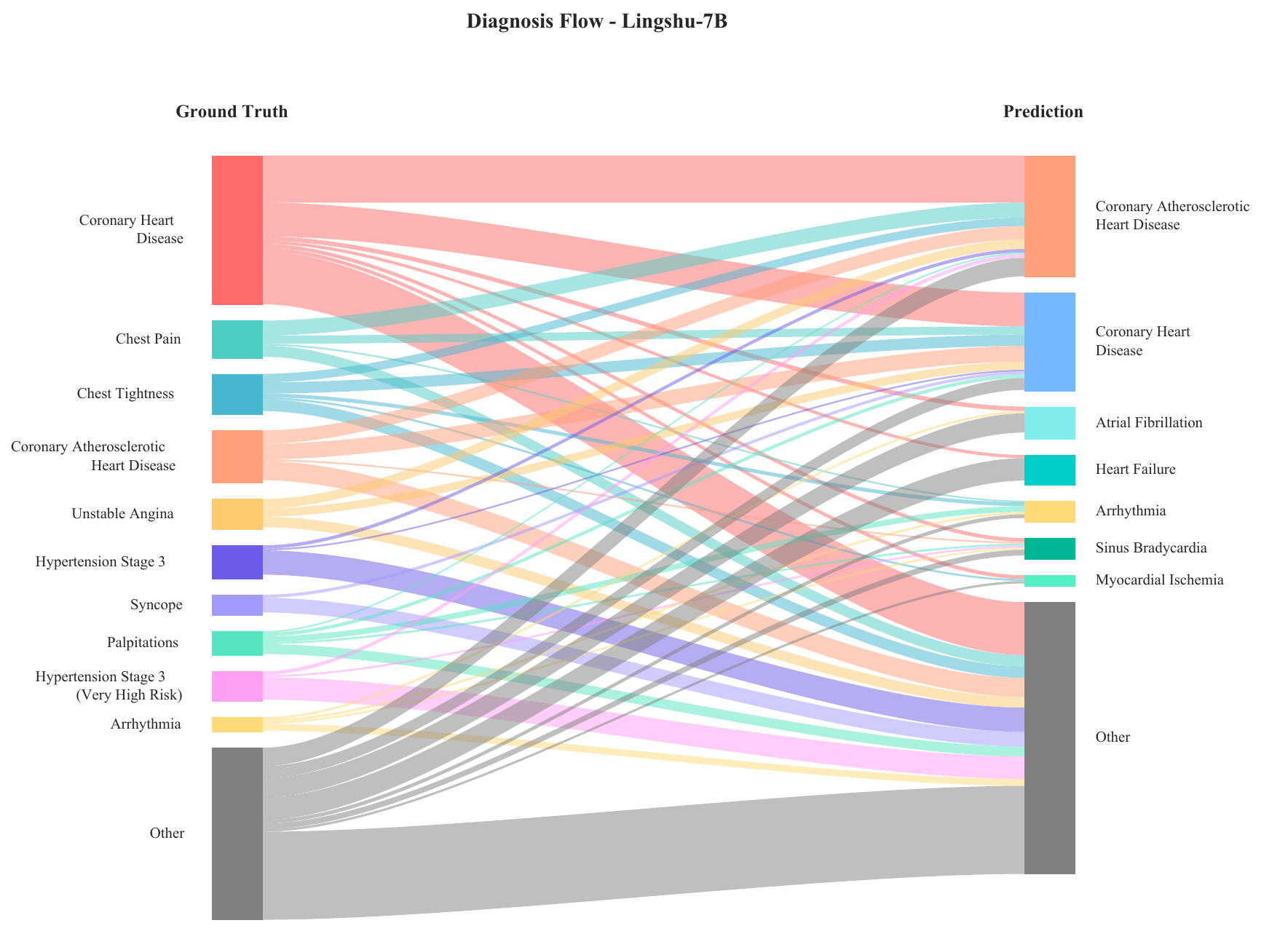}
\hfill
\includegraphics[width=0.32\textwidth]{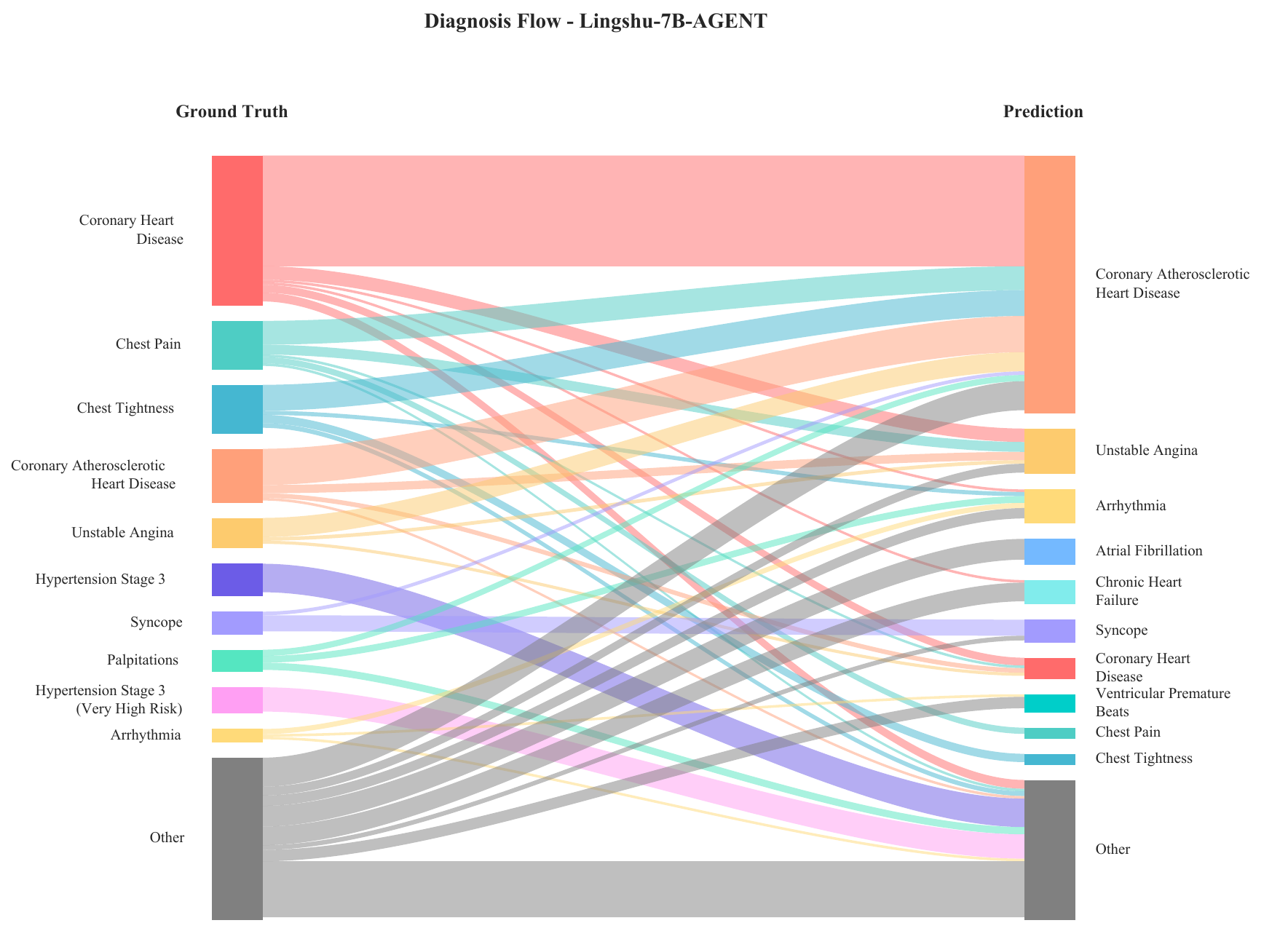}
\caption{\textbf{Horizontal Comparison of Diagnostic Flows (Sankey Diagrams)}. From left to right: Qwen2.5-VL-7B-Instruct, Lingshu-7B, and Lingshu-7B-AGENT. The diagrams visualize the flow from Ground Truth (left side of each plot) to Model Predictions (right side). The Agent model shows thicker, more consistent flows towards correct diagnoses compared to the fragmented flows of the baseline models.}
\label{fig:sankey_diagrams_comparison}
\end{figure}

\addcontentsline{apc}{subsection}{\protect\numberline{\thesubsection}Misdiagnosis patterns and diagnostic flows}
\label{app:errors}
To describe diagnosis-label confusions across models, we conducted a comparative analysis of misdiagnosis patterns for the top 10 diseases using Confusion Matrices (Figure \ref{fig:confusion_matrices_comparison}) and Sankey diagrams (Figure \ref{fig:sankey_diagrams_comparison}). We selected three representative models for horizontal comparison: Qwen2.5-VL-7B-Instruct (General), Lingshu-7B (Medical), and Lingshu-7B-AGENT (Agent).

\noindent \textbf{Pattern Analysis and Safety Risks}. 
Qwen2.5-VL shows more dispersed label confusions, whereas Lingshu-7B has a narrower pattern with persistent confusion between related conditions. The agent shows clearer diagonal alignment in these examples. These descriptive plots do not establish hallucination rates, reasoning correctness, or safety improvements.

\noindent\textbf{Interpretation.} Persistent confusion pairs can guide physician review and further error analysis. Whether retrieved evidence improves clinical decisions or reduces harm requires direct assessment beyond these plots.

\subsection{Additional experimental results}
\addcontentsline{apc}{subsection}{\protect\numberline{\thesubsection}Additional experimental results}
The following figures describe general and medical LLM performance across the reported diagnosis-count groups, disease categories, other subgroups, and symptom-clarity categories. These comparisons are exploratory and do not isolate the effects of model training.

\subsubsection{Diagnosis-count groups}
The diagnosis-count comparison is summarized in Figure~\ref{fig:comorbidity_comparison}. Interpretation depends on the definition and size of each group. Multiple stored reference entries should not be treated as a complete measure of a patient's comorbidities.

\subsubsection{Disease Type Analysis}
\begin{figure}[htbp]
    \centering
    \begin{subfigure}[b]{0.48\textwidth}
        \centering
        \includegraphics[width=\textwidth]{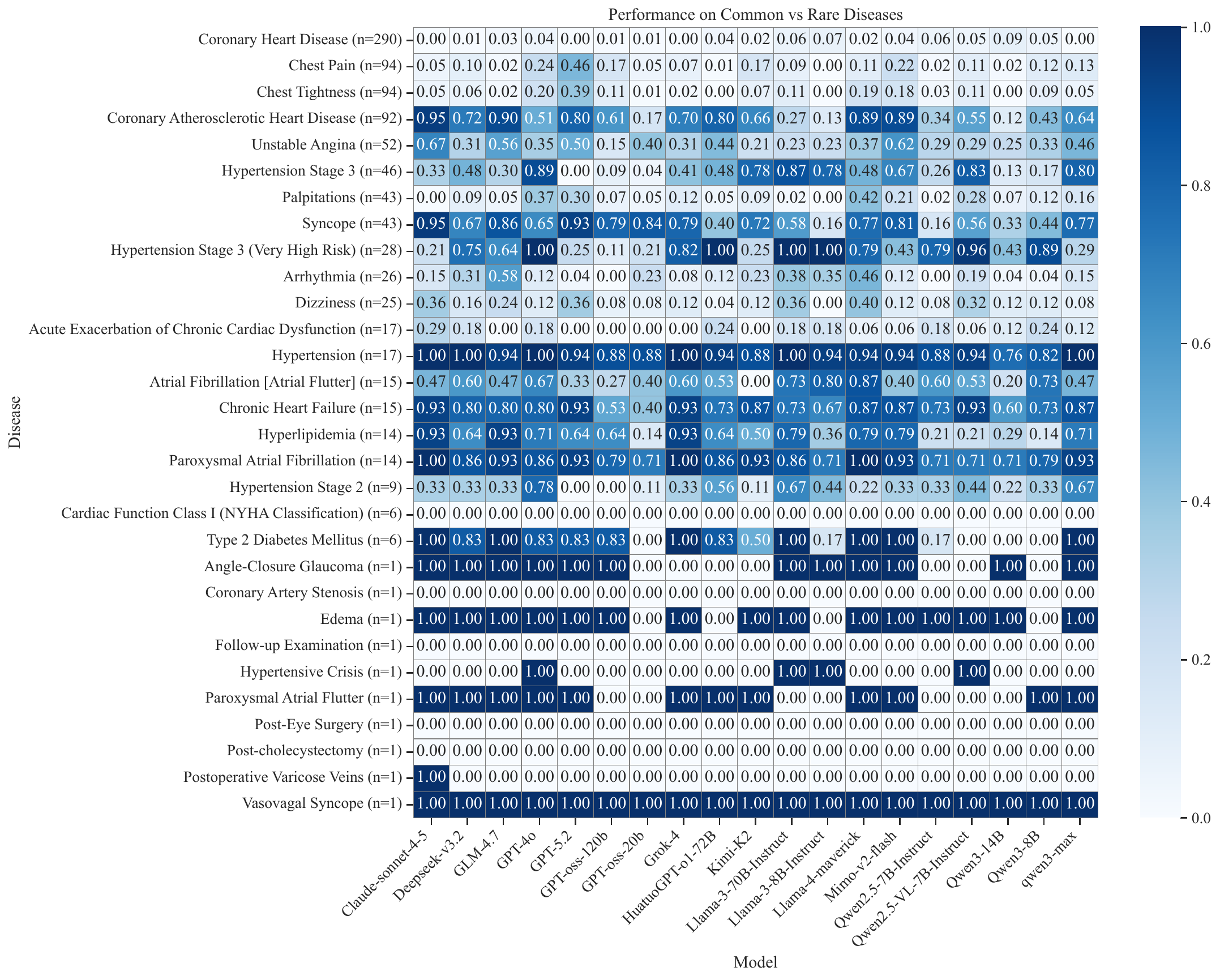}
        \caption{General LLMs}
    \end{subfigure}
    \hfill
    \begin{subfigure}[b]{0.48\textwidth}
        \centering
        \includegraphics[width=\textwidth]{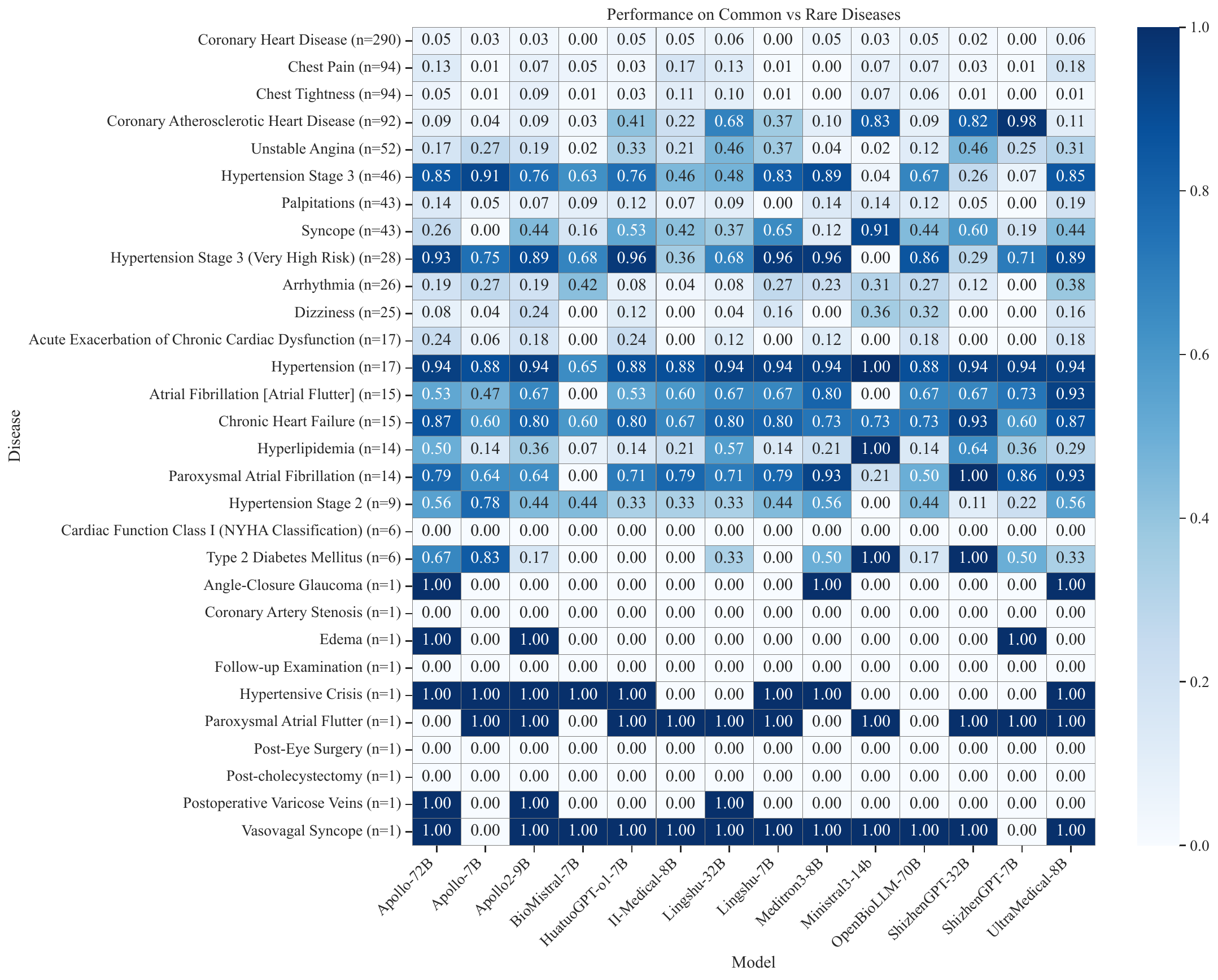}
        \caption{Medical LLMs}
    \end{subfigure}
    \caption{\textbf{Disease Type Performance Heatmap.} These heatmaps display the performance of General and Medical LLMs across various cardiovascular disease types. The color intensity represents the accuracy, allowing for a visual comparison of strengths and weaknesses in diagnosing specific conditions.}
    \label{fig:disease_type_comparison}
\end{figure}

Figure~\ref{fig:disease_type_comparison} compares the reported performance of general and medical LLMs across disease categories. The heatmaps describe category-specific variation and can help identify groups for further error analysis. They do not establish that a model has a deeper understanding of the corresponding diseases.

\subsubsection{Subgroup Capabilities}
\begin{figure}[htbp]
    \centering
    \begin{subfigure}[b]{0.48\textwidth}
        \centering
        \includegraphics[width=\textwidth]{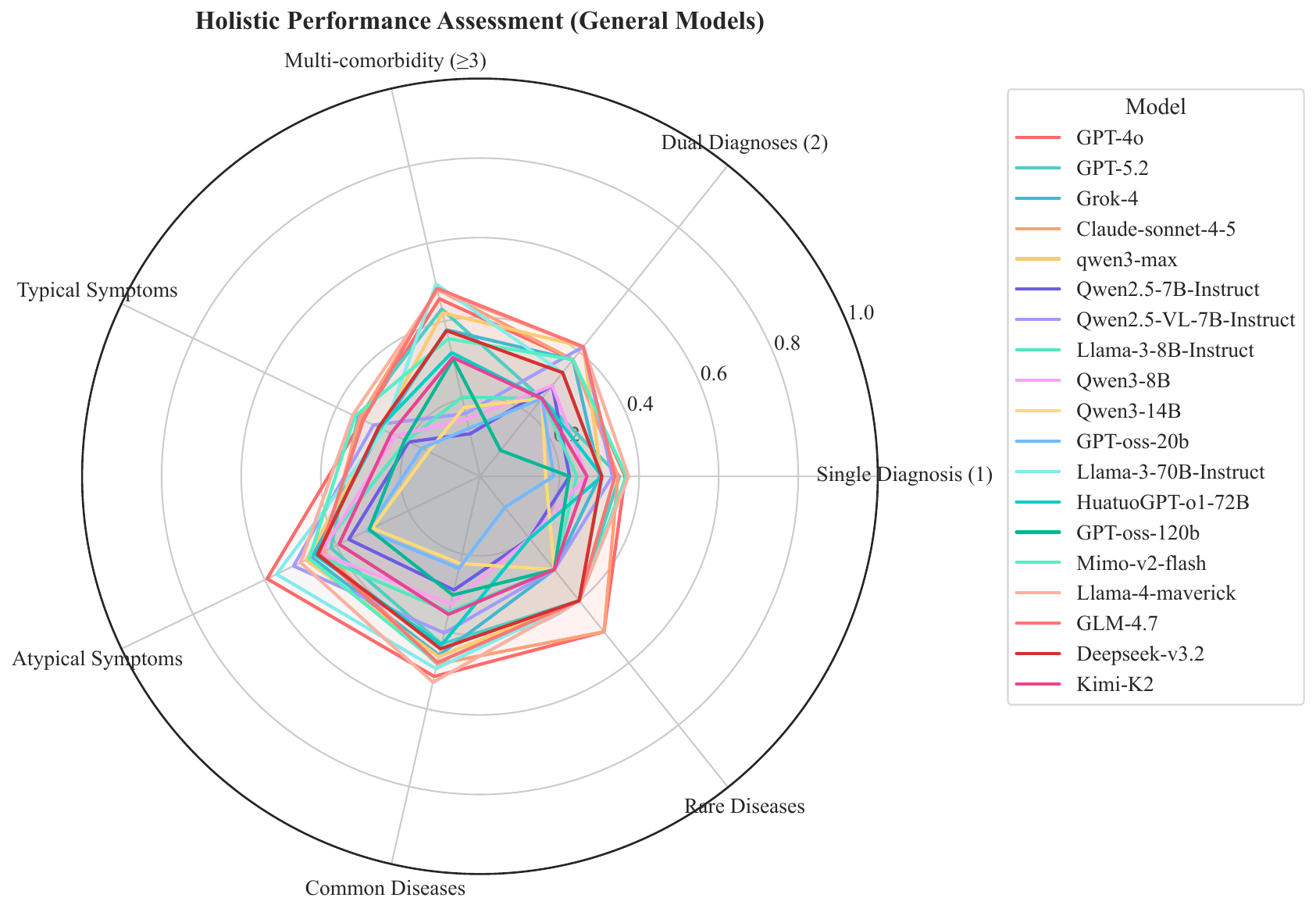}
        \caption{General LLMs}
    \end{subfigure}
    \hfill
    \begin{subfigure}[b]{0.48\textwidth}
        \centering
        \includegraphics[width=\textwidth]{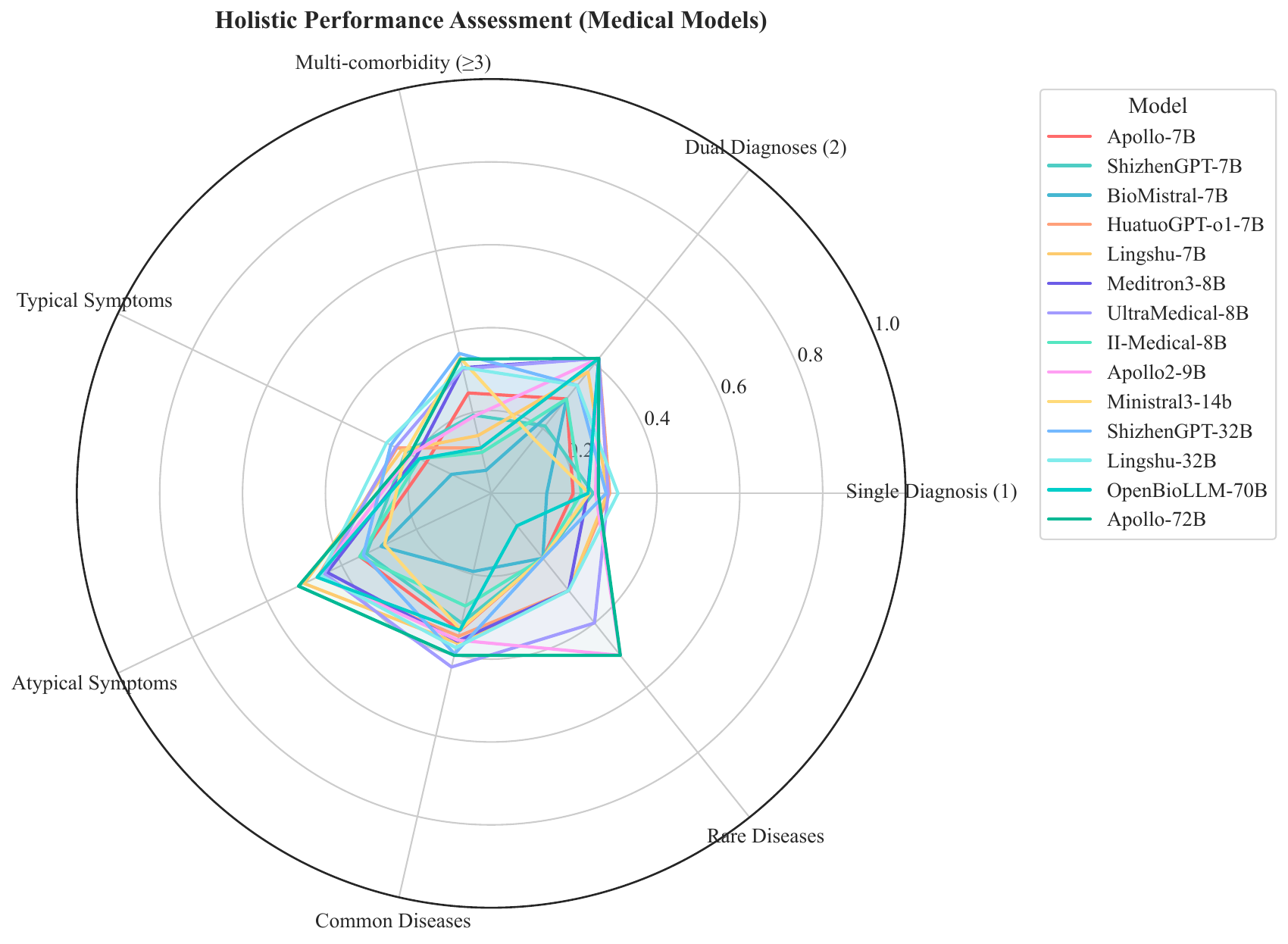}
        \caption{Medical LLMs}
    \end{subfigure}
    \caption{\textbf{Subgroup Performance Radar Chart.} The radar charts provide a multidimensional comparison of General and Medical LLMs across different clinical subgroups. This visualization helps identify specific areas where one model type may outperform the other.}
    \label{fig:subgroup_radar_comparison}
\end{figure}

Figure~\ref{fig:subgroup_radar_comparison} summarizes reported performance across the selected subgroups. The profiles describe variation within this evaluation; establishing fairness or robustness would require explicit subgroup definitions, sample sizes, and uncertainty estimates.

\subsubsection{Symptom Clarity}
\begin{figure}[htbp]
    \centering
    \begin{subfigure}[b]{0.48\textwidth}
        \centering
        \includegraphics[width=\textwidth]{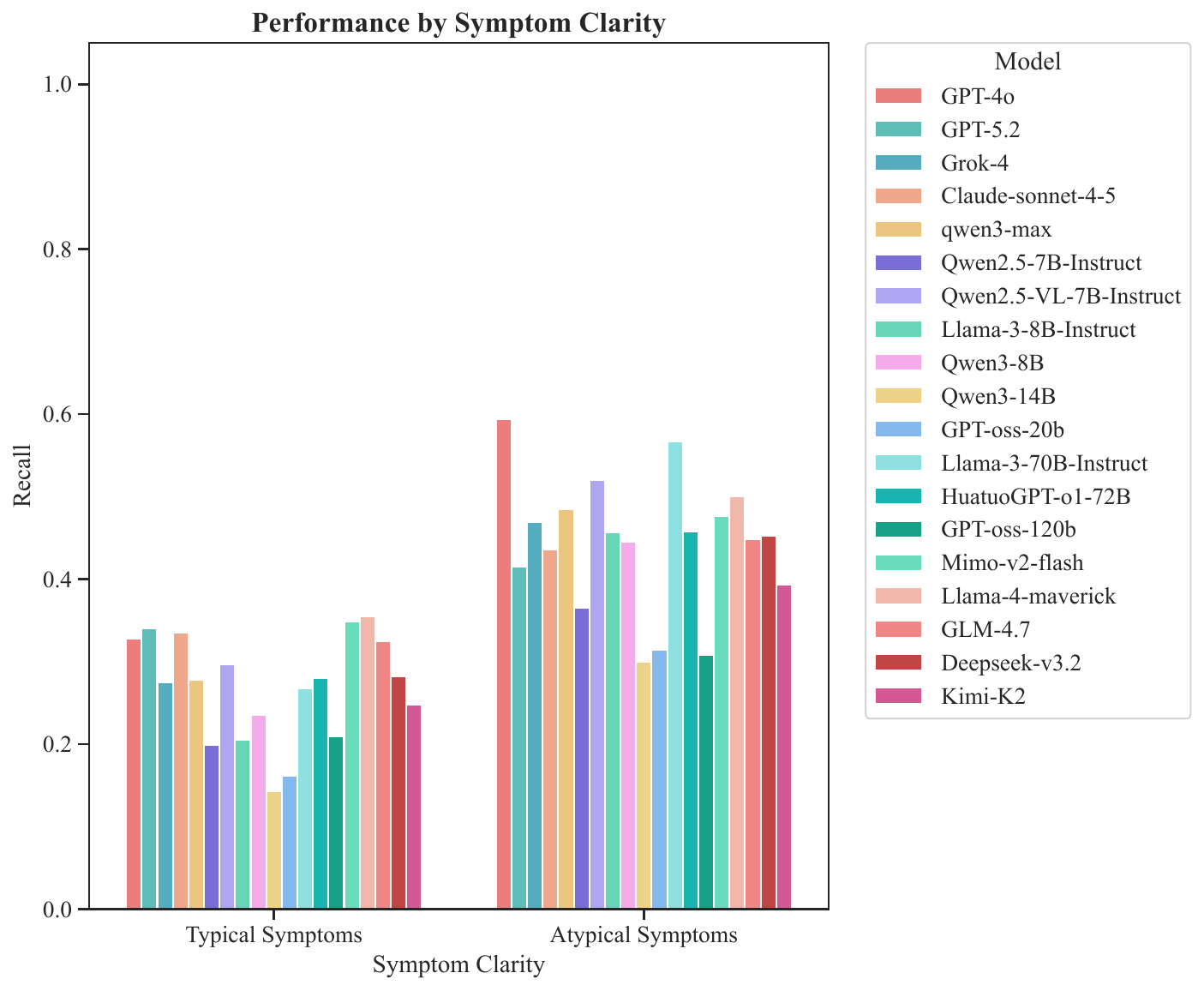}
        \caption{General LLMs}
    \end{subfigure}
    \hfill
    \begin{subfigure}[b]{0.48\textwidth}
        \centering
        \includegraphics[width=\textwidth]{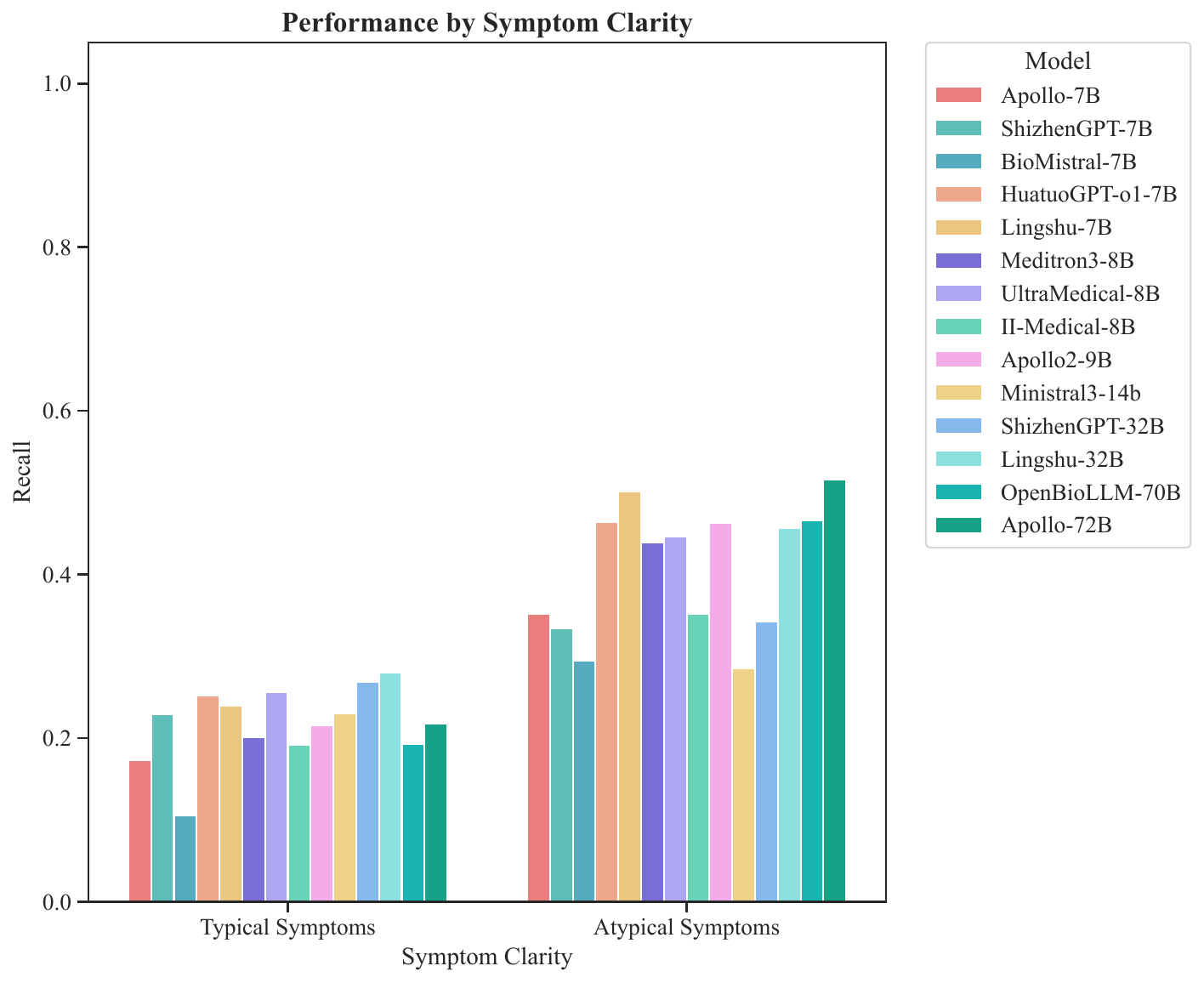}
        \caption{Medical LLMs}
    \end{subfigure}
    \caption{\textbf{Symptom Clarity Performance Comparison.} This figure compares general and medical LLM performance across the symptom-clarity categories in the original analysis.}
    \label{fig:symptom_clarity_comparison}
\end{figure}

Clinical notes vary in the explicitness of symptom descriptions. Figure~\ref{fig:symptom_clarity_comparison} compares reported performance across symptom-clarity categories. Differences between groups are descriptive and do not establish that specialization causes better inference from implicit clinical cues.

\FloatBarrier

\FloatBarrier
\section{Benchmark Prompts}
\addcontentsline{apc}{section}{\protect\numberline{\thesection}Benchmark Prompts}
\label{app:prompts}

To ensure standardized evaluation, we utilized specific prompts for the large language models. The prompts were designed to guide the models in generating structured diagnostic outputs based on the provided clinical notes and ECG results.

\subsection{System Prompt}
\addcontentsline{apc}{subsection}{\protect\numberline{\thesubsection}System Prompt}
The system prompt establishes the role and output format for the model.

\begin{tcolorbox}[benchmarkprompt,title=System Prompt (English)]
``You are a clinical cardiologist responsible for generating a structured final diagnosis based on the provided medical records and ECG results. Please strictly output a JSON object and do not output superfluous text. Output in Chinese, with standardized fields and values.''
\end{tcolorbox}

\begin{tcolorbox}[benchmarkprompt,title=System Prompt (Original Chinese)]
``你是一名心血管内科的临床医生，负责根据给定的病历文书和心电检查结果，生成结构化的最终诊断。请严格输出 JSON 对象，不要输出多余文本。中文输出，字段与取值尽量标准化。''
\end{tcolorbox}

\subsection{User Prompt Template}
\addcontentsline{apc}{subsection}{\protect\numberline{\thesubsection}User Prompt Template}
The user prompt provides the specific case data and reiterates the output requirements.

\begin{tcolorbox}[benchmarkprompt,title=User Prompt Template (English)]
``Task: Final Diagnosis. Input medical records and ECG items, findings, and diagnoses; output the final diagnosis.

Medical Record (Summary): \{medical\_record\}

ECG Results: \{ecg\_results\}

Please return only JSON, containing these keys:
- Diagnosis: Array, where each item is an object containing only the key 'Diagnosis Name' (standardized Chinese primary diagnosis)

Strict Requirements:
1) Return a valid JSON object, with key names and types meeting requirements
2) Do not add explanations or extra fields''
\end{tcolorbox}

\begin{tcolorbox}[benchmarkprompt,title=User Prompt Template (Original Chinese)]
``任务：最终诊断。输入病历文书和心电检查项目、所见、诊断，输出最终诊断。\\
病历文书（摘要）：\{medical\_record\}\\
ECG结果：\{ecg\_results\}\\
请仅返回 JSON，包含这些键：\\
- 诊断: 数组，每项为对象，仅包含键 '诊断名称'（规范中文主诊断）\\
严格要求：\\
1) 返回合法 JSON 对象，键名与类型符合要求\\
2) 不要加入解释或额外字段''
\end{tcolorbox}

\FloatBarrier
\section{Extended Discussion}
\addcontentsline{apc}{section}{\protect\numberline{\thesection}Extended Discussion}
\label{app:discussion}

Primary-diagnosis identification and full-list quality need separate examination. A higher composite score reflects the balance imposed by matching, prediction rank, reference priority, and precision weighting; it is not an established ordering of clinical utility.

\noindent\textbf{Evaluation validation.} Future work should obtain blinded physician ratings of anonymized lists, covering primary-diagnosis correctness, coverage, and inappropriate additions. Agreement with individual metrics and the composite should be assessed on the same cases. Fixed-output-count comparisons and sensitivity analyses for matching thresholds, reference weights, and $\beta$ would help distinguish clinical agreement from length or scoring artifacts. These studies have not been reported here.

\noindent\textbf{Knowledge augmentation.} The 26.40-to-45.80 gain compares SFT with SFT plus the full agent. A follow-up should hold the backbone and approximate token/tool budget fixed across no retrieval, text RAG, graph retrieval, guideline retrieval, and the combined workflow. Present results provide a knowledge-enhanced baseline without isolating graph retrieval's causal contribution.

\noindent\textbf{Input protocol.} Historical records contain diagnostic interpretations of examinations. Timestamped inputs and patient linkage are needed before prospective or patient-independent claims. Removing fields defines a different input protocol and requires new predictions; historical results must not be relabeled as cleaned-input results. Multi-center validation and review of reference ambiguity would further strengthen the benchmark.
\end{document}